\documentclass{article}

\usepackage{microtype}
\usepackage{graphicx}
\usepackage{subfigure}
\usepackage{booktabs} 

\usepackage{hyperref}

\usepackage[accepted]{icml2024}

\usepackage{amsmath}
\usepackage{amssymb}
\usepackage{mathtools}
\usepackage{amsthm}
\usepackage{tikz}
\usetikzlibrary{positioning}
\usepackage{tablefootnote}
\usepackage{tabularx}
\usepackage{threeparttable}
\usepackage{multirow}
\graphicspath{{Images/}}
\usepackage{subfiles}
\usepackage[capitalize,noabbrev]{cleveref}

\theoremstyle{plain}
\newtheorem{theorem}{Theorem}[section]

\newtheorem{corollary}[theorem]{Corollary}
\theoremstyle{definition}
\newtheorem{definition}[theorem]{Definition}

\theoremstyle{remark}

\usepackage[textsize=tiny]{todonotes}

\icmltitlerunning{Fourier-Mixed Window Attention}

\date{}

\begin{document}

\twocolumn[
\icmltitle{Fourier-Mixed Window  Attention: Accelerating Informer \\for Long Sequence Time-Series Forecasting}



\icmlsetsymbol{equal}{*}

\begin{icmlauthorlist}
\icmlauthor{Nhat Thanh Tran}{yyy}
\icmlauthor{Jack Xin}{yyy}
\end{icmlauthorlist}

\icmlaffiliation{yyy}{Department of Mathematics, University of California, Irvine , California, USA}

\icmlcorrespondingauthor{Nhat Thanh Tran}{nhattt@uci.edu}

\icmlkeywords{time-series, window attention}

\vskip 0.3in
]

\printAffiliationsAndNotice{\icmlEqualContribution} 


\begin{abstract}
We study a fast local-global window-based attention method to accelerate Informer for long sequence time-series forecasting.
While window attention being local is a considerable computational saving, it lacks the ability to capture global token information which is compensated by 
 a subsequent Fourier transform block. 
 Our method, named FWin, does not rely on query sparsity hypothesis and an empirical approximation underlying the ProbSparse attention of Informer. Through experiments on univariate and multivariate datasets, we show that 
FWin transformers improve the overall prediction accuracies of Informer while accelerating its inference speeds by 1.6 to 2 times. We also provide a mathematical definition of FWin attention, and prove that it is equivalent to the canonical full attention under the block diagonal invertibility (BDI) condition of the attention matrix. The BDI is shown experimentally to hold with high probability for typical benchmark datasets.

\end{abstract}


\section{Introduction}

Recent progress in long sequence time-series forecasting (LSTF)  
has been led by either transformers with sparse attention (\cite{informer_21} and references therein) or attention in combination with signal pre-processing such as seasonal-trend decomposition \cite{fedformer}
or adopting auto-correlation to account for periodicity in the data \cite{autoformer_21}. On the other hand, Fourier transform has been proposed as an alternative mixing tool in lieu of standard attention \cite{vaswani2017attention} to speed up prediction in natural language processing (NLP) tasks (FNet, \cite{Fnet}). Though Fourier transform is meant to mimic the mixing functions of multi-layer perceptron (MLP, \cite{mlpmixer_21}), it is not well-understood why it works and when assistance from attention layers remain necessary to maintain performance. 
In computer vision (CV), Fourier transform is also used as a filtering 
step in early stages of transformer (GFNet,\cite{GFN_21}) to enhance 
a fully attention-based architecture. 
A recent advance in CV is to adopt window attention to 
reduce quadratic complexity of full attention \cite{vaswani2017attention}. In shifted window attention (Swin \cite{Swin}), the attention is first computed on non-overlapping windows as a local approximation, then on shifted window configurations to spread local attention globally in the image domain. This local-global approximation of full attention occurs entirely in the image domain, and is repeated over multiple stages in the network. The advantage is that the recipe is independent of the data distribution. In contrast, the ProbSparse self-attention of Informer \cite{ informer_21,haoyietal-informerEx-2023} relies on long tail data distribution to select the top few queries.

We are interested in developing an efficient window-based attention to replace ProbSparse attention and accelerate Informer with no prior knowledge of data properties such as periodicity 
(seasonality) so that our method is applicable in a general context of time series. We also refrain from pre-processing 
data to increase performance as this step can be added later. The main issue is how to globalize the local window 
attention without performing shifts, 
since Informer only has two attention blocks in the encoder and is unable to 
facilitate repeated window shifting as in a deeper network Swin \cite{Swin}.


We propose to replace ProbSparse attention of Informer via a (local) window attention followed by a Fourier transform (mixing) layer, a novel local-global attention which we call Fourier-Mixed window attention (FWin). The resulting network, FWin Transformer, aims to reduce the  complexity of the full attention \cite{vaswani2017attention} and approximate its functionality by mixing the window attention. Instead of shifting windows for globalization \cite{Swin}, we employ the parameter free fast Fourier Transform (FFT) to generate connections among the tokens. The strategy allows the window attention layer to focus on learning local information, while the Fourier layer effectively mixes tokens and spreads information globally. In the ablation study, we find that the network prediction accuracy is lower if we replace ProbSparse by only Fourier mixing as in FNet \cite{Fnet} without the help of window attention.
Besides conducting extensive experiments on FWin to support our methodology, we also provide a mathematical formulation of mix window attention and prove that it is equivalent to the canonical attention.




Our main contributions in this paper are summarized below. 
\begin{itemize}

\item We propose FWin and replace the ProbSparse self attention block (Fig. \ref{fig:model_overview}, left) via a 
window self-attention block followed by a Fourier mixing layer (Fig. \ref{fig:model_overview}, right) in both the encoder and decoder of Informer \cite{informer_21, haoyietal-informerEx-2023}. 

\item We show experimentally that FWin either increases or maintains Informer's performance level while significantly accelerating its inference speed (by about 1.6 to 2 times) on both univariate and multivariate LSTF data. The training times of FWin are consistently lower than those of Informer across various data. Inference speeds of FWin exceed those of Fedformer \cite{fedformer} and Autoformer \cite{autoformer_21} by a factor of 5 to 10 with shorter training times overall. 

\item We propose FWin-S, a light weight version of FWin, by removing Fourier mixing layer in the decoder (Fig. \ref{fig:model_overview}, right); and present its competitive performance with faster inference speed.


\item We provide a mathematical formulation of mix window attention and prove that it is equivalent to the canonical attention under under block diagonal invertibility (BDI) condition of the attention matrix.
\end{itemize}

{\bf Organization.} This paper will proceed as follows: In section 2, we provide background on full attention, window attention, Fourier mixing.
In Section 3, we present
FWin methodology and its complexity. In section 4, we 
present experimental results and analysis. In section 5, 
we mathematically formulate mix window attention and prove it is equivalent to canonical attention. We show that FWin is a special case of mix window attention. 
The paper ends with concluding remarks. Related works, ablation studies, FWin in  a nonlinear/non-parametric  regression model, and more experimental details are provided in the Appendix.



\section{Background and Preliminary}

{\bf Canonical Full Attention}
Given an input sequence $x\in\mathbb{R}^{L\times d_{\text{model}}}$, where $L$ is the sequence length and $d_{\text{model}}$ is the embedded dimension of the model. The input $x$ is then converted into queries ($Q$), keys ($K$), values ($V$) as:
 \[   Q = xW_Q + b_Q,\,
    K = xW_K + b_K,\,
    V = xW_V + b_V,\]
where $W_Q, W_K, W_V\in \mathbb{R}^{d_{\text{model}}\times d_{\text{attn}}}$ are the weighted matrix, and $b_Q, b_K, b_V\in\mathbb{R}^{L\times d_{\text{attn}}}$ are the bias matrix. In most cases, we will have $d_{\text{model}}=d_{\text{attn}}$, thus we will refer to $d_{\text{model}}$ for the remaining of the paper. 

We have
\begin{equation}\label{eqn:attn_eqn}
     Attn_f(Q, K, V) = \text{softmax}(QK^T/\sqrt{d_{\text{model}}}) V,
\end{equation}
where $Attn_f(\cdot)$ is the attention function.

We refer to the function 
in (\ref{eqn:attn_eqn})
as full attention \cite{vaswani2017attention}, because it involves the interaction of all the key and query pairs. The final output is the weighted sum of all the values.

{\bf Window Attention}
The full attention computation involves the dot product between each query and all the keys. However, for tasks with large sequence lengths such as processing of high resolution images, the computational cost of full attention can be significant \cite{Swin}. As in Swin Transformer, we divide the sequence into subsequences of smaller length, compute sub-attention for each of the subsequences individually and then concatenate all the sub-attention together. Namely, we divide sequence $x$ into $N$ subsequences: $x^{(1)}, x^{(2)},\dots, x^{(N)}$, such that $x = [x^{(1)}, x^{(2)},\dots, x^{(N)}]^T$. Each $x^{(i)}\in \mathbb{R}^{L/N\times d_{\text{model}}}$ for $i=1,2,\dots, N$, where $N=L/w$, $w$ is a fixed window size. This implies we divide the queries, keys and values as follow $Q = [Q^{(1)}, Q^{(2)},\dots, Q^{(N)}]^T$, $K = [K^{(1)}, K^{(2)},\dots, K^{(N)}]^T$, $V = [V^{(1)}, V^{(2)},\dots, V^{(N)}]^T$. Thus we compute attention for each subsequence as follows: 
\begin{equation}
    Attn_f(Q^{(i)},K^{(i)},V^{(i)}) = \text{softmax}(\,\dfrac{Q^{(i)}K^{(i)T}}{\sqrt{d_{\text{model}}}}\, ) V^{(i)}.
\end{equation}
After computing the attention for each subsequence, we concatenate the sub attentions to form the window attention:
\begin{equation}
Attn_w(Q,K,V) = 
\begin{bmatrix}
    Attn_f(Q^{(1)},K^{(1)},V^{(1)}) \\
    \vdots\\ 
    Attn_f(Q^{(N)},K^{(N)},V^{(N)})
\end{bmatrix}.    
\end{equation}
In window attention, we compute attention on a window-by-window basis. Within each window, all the keys are multiplied with the corresponding query within that window. The output is the weighted sum of the values within the same window, rather than considering the entire set of values. This approach reduces the computational cost of computing attention on a sequence of length $L$ from $\mathcal{O}(L^2)$ 
of full attention to 
$\mathcal{O}(L w)$, where $w$ is a fixed window size.

A natural question that arises is how well the window attention compares to full attention. If softmax$(QK^T/\sqrt{d_{\text{model}}})$ is almost diagonal, meaning that most of the entries apart from the diagonal are close to zero, then window attention can be considered a good approximation of full attention. However, this is often not the case as seen in a nonlinear regression model in the Appendix \ref{sec:nonparam}.

Window attention restricts the interaction between queries and keys by only allowing queries to attend to their local window keys. As a result, window attention provides limited or local information. On the other hand, full attention enables queries to attend to keys that are further away, allowing for  global interaction. If we replace full attention with window attention, our model may lack a comprehensive understanding of global information. Therefore, it is desirable for our model to retain some level of global information when using window attention as a substitute for full attention.
To incorporate global information, Swin Transformer \cite{Swin} introduces shifted window attention. In this approach, before dividing the input sequence $x$ into sub-sequences, one performs a circular shift of the indices of $x$ by certain value. This shift allows the ending values of $x$ to become the starting values of our input. The circular shifting process is repeated for each consecutive window attention layers.

{\bf FNet}
Another way to promote global token interaction is by Fourier transform as proposed in FNet \cite{Fnet}.
Given input $x\in \mathbb{R}^{L\times d_{\text{model}}}$, one computes Fourier transform along the model dimension ($d_{\text{model}}$),  then along the time dimension ($L$), finally taking real part to arrive at:
\begin{equation}
    y = \mathcal{R}(\mathcal{F}_{\text{time}}(\mathcal{F}_{\text{model}}(x))),
\end{equation}
where $\mathcal{F}$ is 1D discrete Fourier transform (DFT), and $\mathcal{R}$ is the real part of a complex number.
Since DFT is free of learning parameter, 
one would eventually pass the transformed sequence
through a Feed Forward  fully-connected layer (FC). This approach can be interpreted as the Fourier transform being an effective mechanism for mixing sub-sequences (tokens) \cite{Fnet}. By applying the Fourier transform, the Feed Forward layer gains access to all the tokens. 

\section{Methodology}

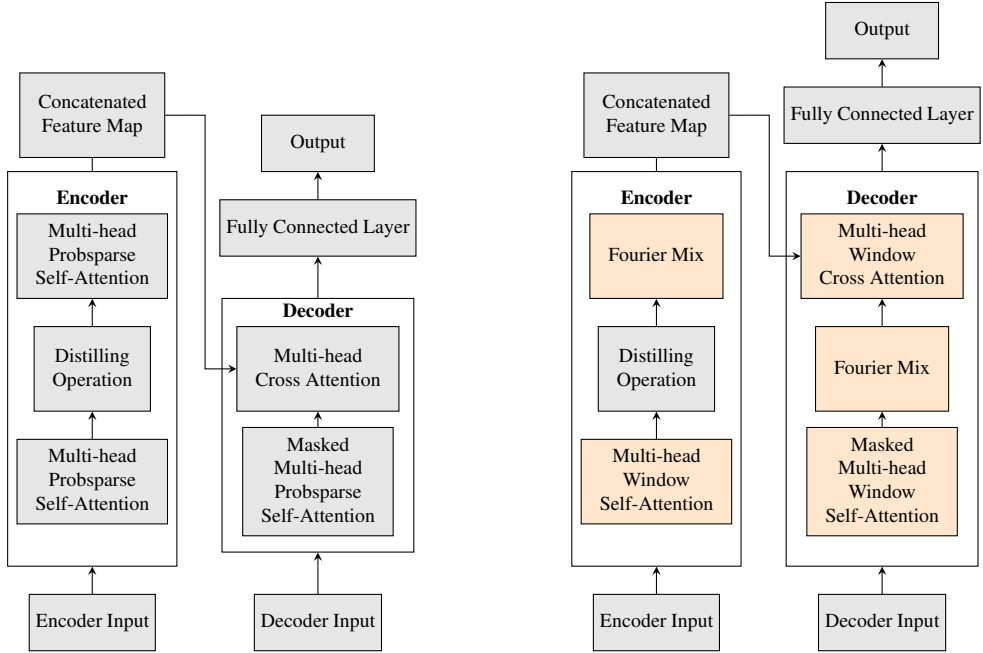
\begin{figure*}[ht]
    \centering

\begin{tikzpicture}[scale=0.75,every node/.style={transform shape}, thin]





\node[rectangle, draw,minimum width=2cm, minimum height=1cm, fill=gray!20] at (0.5,-1.5) (encin){Encoder Input};

\node[rectangle, draw,minimum width=2cm, minimum height=1.5cm, fill=gray!20] at (0.5,1) (encattn){\begin{tabular}{c}
     Multi-head\\
     Probsparse\\
     Self-Attention
\end{tabular}};

\node[rectangle, draw,minimum width=2cm, minimum height=1.5cm, fill=gray!20] at (0.5,3) (encdistil){\begin{tabular}{c}
     Distilling  \\
     Operation
\end{tabular}};

\node[rectangle, draw,minimum width=2cm, minimum height=1.5cm, fill=gray!20] at (0.5,5) (encfnet){\begin{tabular}{c}
     Multi-head\\
     Probsparse\\
     Self-Attention
\end{tabular}};

\node[minimum width=2cm, minimum height=1.5cm] at (0.5,6) (encoder){\begin{tabular}{c}
     \textbf{Encoder}
\end{tabular}};

\node[rectangle, draw,minimum width=3cm, minimum height=7cm] at (0.5,3) (encblock){};

\node[rectangle, draw,minimum width=2cm, minimum height=1.5cm, fill=gray!20] at (0.5,7.5) (encfeat){\begin{tabular}{c}Concatenated \\Feature Map \end{tabular}};


\node[rectangle, draw,minimum width=2cm, minimum height=1cm, fill=gray!20] at (4.5,-1.5) (decin){Decoder Input};

\node[rectangle, draw,minimum width=2cm, minimum height=1.5cm, fill=gray!20] at (4.5,1) (decsattn){\begin{tabular}{c}
     Masked \\Multi-head  \\
     Probsparse\\
     Self-Attention
\end{tabular}};

\node[rectangle, draw,minimum width=2cm, minimum height=1.5cm, fill=gray!20] at (4.5,3) (deccattn){\begin{tabular}{c}
     Multi-head  \\
     Cross Attention
\end{tabular}};

\node[minimum width=2cm, minimum height=1.5cm] at (4.5,4) (encoder){\begin{tabular}{c}
     \textbf{Decoder}
\end{tabular}};

\node[rectangle, draw,minimum width=3.4cm, minimum height=4.5cm] at (4.5,2) (decblock){};

\node[rectangle, draw,minimum width=2cm, minimum height=1cm, fill=gray!20] at (4.5,5.5) (decfcl){Fully Connected Layer};

\node[rectangle, draw,minimum width=2cm, minimum height=1cm, fill=gray!20] at (4.5,7.0) (decout){Output};

\draw[-stealth] (encin) -- (encblock);
\draw[-stealth,scale=0.5] (encattn) -- (encdistil);
\draw[-stealth,scale=0.5] (encdistil) -- (encfnet);
\draw[-] (encblock) -- (encfeat);
\draw[-stealth,scale=0.5] (decin) -- (decblock);
\draw[-stealth,scale=0.5] (decsattn) -- (deccattn);
\draw[-stealth,scale=0.5] (decblock) -- (decfcl);
\draw[-stealth,scale=0.5] (decfcl) -- (decout);
\draw[-stealth] (encfeat) --(2.5,7.5)--(2.5,3)-- (deccattn);



\node[rectangle, draw,minimum width=2cm, minimum height=1cm, fill=gray!20] at (10.5,-1.5) (encin){Encoder Input};

\node[rectangle, draw,minimum width=2cm, minimum height=1.5cm, fill=orange!20] at (10.5,1) (encattn){\begin{tabular}{c}
     Multi-head  \\
     Window\\
     Self-Attention
\end{tabular}};

\node[rectangle, draw,minimum width=2cm, minimum height=1.5cm, fill=gray!20] at (10.5,3) (encdistil){\begin{tabular}{c}
     Distilling  \\
     Operation
\end{tabular}};

\node[rectangle, draw,minimum width=2cm, minimum height=1.5cm, fill=orange!20] at (10.5,5) (encfnet){\begin{tabular}{c}
     Fourier Mix
\end{tabular}};

\node[minimum width=2cm, minimum height=1.5cm] at (10.5,6) (encoder){\begin{tabular}{c}
     \textbf{Encoder}
\end{tabular}};

\node[rectangle, draw,minimum width=3cm, minimum height=7cm] at (10.5,3) (encblock){};

\node[rectangle, draw,minimum width=2cm, minimum height=1.5cm, fill=gray!20] at (10.5,7.5) (encfeat){\begin{tabular}{c}Concatenated \\Feature Map \end{tabular}};


\node[rectangle, draw,minimum width=2cm, minimum height=1cm, fill=gray!20] at (14.5,-1.5) (decin){Decoder Input};

\node[rectangle, draw,minimum width=2cm, minimum height=1.5cm, fill=orange!20] at (14.5,1) (decsattn){\begin{tabular}{c}
     Masked \\Multi-head  \\
     Window\\
     Self-Attention
\end{tabular}};

\node[rectangle, draw,minimum width=2cm, minimum height=1.5cm, fill=orange!20] at (14.5,3) (decfmix){\begin{tabular}{c}
     Fourier Mix
\end{tabular}};

\node[rectangle, draw,minimum width=2cm, minimum height=1.5cm, fill=orange!20] at (14.5,5) (deccattn){\begin{tabular}{c}
     Multi-head  \\
     Window\\
     Cross Attention
\end{tabular}};

\node[minimum width=2cm, minimum height=1.5cm] at (14.5,6) (encoder){\begin{tabular}{c}
     \textbf{Decoder}
\end{tabular}};

\node[rectangle, draw,minimum width=3.4cm, minimum height=7cm] at (14.5,3) (decblock){};

\node[rectangle, draw,minimum width=2cm, minimum height=1cm, fill=gray!20] at (14.5,7.5) (decfcl){Fully Connected Layer};

\node[rectangle, draw,minimum width=2cm, minimum height=1cm, fill=gray!20] at (14.5,9.0) (decout){Output};

\draw[-stealth] (encin) -- (encblock);
\draw[-stealth,scale=0.5] (encattn) -- (encdistil);
\draw[-stealth,scale=0.5] (encdistil) -- (encfnet);
\draw[-] (encblock) -- (encfeat);
\draw[-stealth,scale=0.5] (decin) -- (decblock);
\draw[-stealth,scale=0.5] (decsattn) -- (decfmix);
\draw[-stealth,scale=0.5] (decfmix) -- (deccattn);
\draw[-stealth,scale=0.5] (decblock) -- (decfcl);
\draw[-stealth,scale=0.5] (decfcl) -- (decout);

\draw[-stealth] (encfeat) --(12.5,7.5)--(12.5,5)-- (deccattn);


\end{tikzpicture}
\caption{Model comparison: Informer (left), FWin (right, orange color denotes our contributions); FWin-S (FWin with its decoder's Fourier Mix block removed).}
\label{fig:model_overview}
\end{figure*}

{\bf Informer Overview}
The input $x\in\mathbb{R}^{L\times d_{\textbf{data}}}$ passes through an embedding layer to encode the time scale information and return $X\in \mathbb{R}^{L\times d_{\text{model}}}$. In the encoder, each layer consists of an attention block followed by a distilling convolution with MaxPool of stride 2 and a down-sampling to halve dimension. Thus, with 2 encoder layers, the time dimension of the first attention block is $L$, while the second block input is $L/2$. For both efficiency and causality, the decoder attention has a masked multi-head ProbSparse self-attention structure, see Fig. \ref{fig:model_overview} (left) for an overview. 
The ProbSparse self-attention \cite{informer_21} relies on a sparse query measurement function (an analytical approximation of Kullback-Leibler divergence) so that each key attends to only a few top queries for computational savings. The sparse query hypothesis or equivalently the long tail distribution of self-attention feature map is based on softmax scores in self-attention of a 4-layer canonical transformer trained on ETTh1 data set (Appendix C, \cite{informer_21}). 

{\bf Our Approach }
We introduce Fourier mixed window attention (FWin) to the self-attention blocks in the encoder and decoder of Informer. Specifically, we replace the ProbSparse self-attention blocks in the encoder and decoder with window attention followed by a Fourier mixing layer. 
We also replace multi-head cross-attention in the decoder by a window multi-head cross-attention.
Fig. \ref{fig:model_overview} (right)  illustrates the key components of FWin Transformer. {\it Different from ProbSparse attention, our FWin approach does not rely on whether the query sparse hypothesis holds on a data set}. 

We remark that our model adopted a modified version of the FNet architecture where 
Fourier transform is applied to the input along the model and the time dimensions without a subsequent Feed Forward layer. Partly this is due to the Feed Forward layer already present in the decoder of Informer 
before the output (Fig. \ref{fig:model_overview}, left). We denote this specific component as \textbf{Fourier Mix} in our proposed FWin architecture, as depicted in Fig. \ref{fig:model_overview}, right frame. If the Fourier Mix is removed from the decoder, we have a lighter model called 
FWin-S, which turns out to be a competitive 
design as well (see section 4).

{\bf Motivation }
Our model aims to capture both the short and long variations in the sequential data. The high frequency part  represents local fluctuation, while the low frequency part describes the overall trend of the data. To achieve this, we utilize a combination of window attention and Fourier transform.

The window attention mechanism limits interactions among tokens, enabling the attention layer to learn the local time fluctuations. To capture the overall trend, the tokens need to be mixed beyond the window boundaries. This is achieved by leveraging the Fourier transform. 

Another interpretation of the Fourier Mix layer is that the Fourier transform in the encoder maps the time domain to the frequency domain. As a result, the cross attention layer in the decoder must operate on the frequency domain. To achieve this, the model applies the Fourier transform to the decoder's input before passing it to the cross attention layer. Subsequently, the FC layer maps the output of the cross attention from frequency domain back to the time domain. 

By combining window attention (local) and  Fourier transform (global), our model gains the ability to capture both the local time fluctuations and overall trend of the data. This comprehensive approach ensures that our model can effectively capture the complex dynamics present in the sequential data.

{\bf Encoder }
Each encoder layer is defined as either a window attention layer or a Fourier Mix layer. The layers are interwoven, with the first layer being a window attention. Subsequent layers are connected by a distillation operation. For example, a 3-layer encoder will consist of a window attention layer, a distillation operation, and a Fourier Mix layer, another distillation operation, and finally another window attention layer. Fig. \ref{fig:model_overview} illustrates an encoder with 2 layers. 

{\bf Decoder }
In the decoder, a layer composed of a masked window self-attention  followed by a Fourier Mix and then window cross attention with layer normalization in between. Towards the end of the layer, convolutions and layer normalization are applied. Fig. \ref{fig:model_overview} shows a decoder with one layer. 

{\bf Window Cross Attention }
In a self-attention layer, the query and key vectors have the same time dimension, allowing us to use the same window size to split these vectors. However, in the case of cross attention, this may not hold true, especially if the encoder includes dimension reduction layers. In such cases, the key and value vectors may have a smaller time dimension compared to the query vectors, which originate from the decoder. Additionally, the encoder and decoder inputs may have different input time dimensions, as is the case in our specific problem.
To ensure equal number of attention windows in cross attention, we divide the query, key, and value vectors based on the number of windows rather than the window size. This adjustment accounts for varying time dimensions and guarantees a consistent number of attention windows for the cross attention operation. 

{\bf Complexity}
With the replacements in the attention computation, our approach offers significant complexity reduction compared to Informer. In the encoder, the first attention layer partitions the time dimension of the input by a window of size $w$, by default $w=24$, resulting in each window attention input having dimensions of $w\times d_{\text{model}}$. Thus the cost of this layer is $\mathcal{O}(Lwd_{\text{model}})$. Furthermore, in the second attention layer, the computation of attention is completely replaced by the Fourier Mix layer, eliminating three linear projections for query, key, and value vectors. 
Since we apply the FFT over the time dimension and the model dimension, the total cost for the Fourier Mix layer is $\mathcal{O}(Ld_{\text{model}}\log(Ld_{\text{model}}))$. 

Tab. \ref{tab:Layer_Complexity} is a summary of computational costs for each type of layers we discussed. Tab. \ref{tab:Model_Complexity} compares the complexities of Informer and FWin. 
The $L^2 d_{\text{model}}$ cost of Informer comes from 
full cross attention in its decoder (Fig. \ref{fig:model_overview}, left). In FWin, 
this cost is reduced to 
$L w d_{\text{model}}$ by window cross attention 
(Fig. \ref{fig:model_overview}, right).

\begin{table}[h]
    \caption{Computational complexities of major layers in Informer \cite{informer_21} and FWin transformer.}
    \label{tab:Layer_Complexity}
    \vskip 0.1in
    \centering
\begin{tabular}{|c|c|}
\hline
     
     Full Attention&$\mathcal{O}(L^2d_{\text{model}})$\\
     \hline
     ProbSparse Attention&$\mathcal{O}(L\log(L)d_{\text{model}})$\\
     \hline
     Window Attention&$\mathcal{O}(Lw d_{\text{model}})$ \\
     \hline
     Fourier Mix &$\mathcal{O}(Ld_{\text{model}}\log(Ld_{\text{model}}))$\\
     \hline
\end{tabular}
    \vskip -0.1in
\end{table}

\begin{table}[h]
    \caption{Computational complexity comparison: Informer vs. FWin as shown in Fig. \ref{fig:model_overview}.}
    \vskip 0.1in
    \centering
\begin{tabular}{l|l}

     Informer& $\mathcal{O}(L\log(L)d_{\text{model}}+ L\log(L)d_{\text{model}}$  \\&$+ L\log(L)d_{\text{model}} + L^2d_{\text{model}})$ \\
     \hline
     FWin& $\mathcal{O}(Lw d_{\text{model}} + Ld_{\text{model}}\log(Ld_{\text{model}})$ \\
     &$ + Lw d_{\text{model}} + Ld_{\text{model}}\log(Ld_{\text{model}})$\\
     &$+Lw d_{\text{model}})$  \\

\end{tabular}

    \label{tab:Model_Complexity}
    \vskip -0.1in
\end{table}

\section{Experiment}

 {\bf Datasets}
We experiment on the following public benchmark datasets (additional ones are  described in the Appendix \ref{sec:exp_details}).

\textbf{ETT} (Electricity Transformer Temperature)\footnote{https://github.com/zhouhaoyi/ETDataset}: The dataset contains information of six power load features and target value ``oil temperature". We used two hour datasets ETTh$_1$ and ETTh$_2$, and the minute level dataset ETTm$_1$. The train/val/test split ratio is 6:2:2.


\textbf{Weather} (Local Climatological Data)\footnote{https://www.ncei.noaa.gov/data/local-climatological-data}: The dataset contains local climatological data collected hourly in 1600 U.S. locations over 4 years from 2010 to 2013. The data consists of 11 climate features and target value ``wet bulb". The train/val/test split ratio is 7:1:2.


\textbf{ECL} (Electricity Consuming Load) \footnote{https://archive.ics.uci.edu/ml/ datasets/ElectricityLoadDiagrams20112014.}: This dataset contains electricity consumption (Kwh) of 321 clients. The data convert into hourly consumption of 2 year and ``MT\_320" is the target value. The train/val/test split ratio is 7:1:2.

{\bf Setup}
The default setup of the model parameters is shown in Tab. \ref{tab:Model_hyperparameter}. For Informer, we used their up to date code at: \url{https://github.com/zhouhaoyi/Informer2020}, which incorporated all the functionalities described recently in \cite{haoyietal-informerEx-2023}. In our experiments, we average over 5 runs. The total number of epochs is 6 with early stopping. We optimized the model with Adam optimizer, and the learning rate starts at $1e^{-4}$, decaying two times smaller every epoch. For fair comparison, all of the hyper-parameters are the same across all the models which were trained/tested on a desktop machine with four Nvidia GeForce 8G GPUs. 
Our source code is available upon request.
\begin{table}[t]
    \caption{Model default parameters.}
    \vskip 0.1in
    \centering
    \begin{tabular}{c c|c c}
        $d_{\text{model}}$ & 512  & Window size & 24\\
        $d_{ff}$& 2048& Cross Attn Window no. & 4 \\
        n\_heads & 8 & Epoch & 6\\
        Dropout & 0.05& Early Stopping Counter & 3\\
        Batch Size& 32& Initial Lr & 1e-4\\
        Enc.Layer no.& 2 & Dec.Layer no.& 1\\
    \end{tabular}

    \label{tab:Model_hyperparameter}
    \vskip -0.1in
\end{table}

\subsection{Results and Analysis}

We present a summary of the univariate and multivariate evaluation results for all methods on 7 benchmark datasets in Tab. \ref{tab:Model_accuracy}. MAE $=\frac{1}{n}\sum_{i=1}^n |y- \hat{y}|$  and MSE $=\frac{1}{n}\sum_{i=1}^n (y- \hat{y})^2$ are used as evaluation metrics. The best results are highlighted in boldface, and the total count at the bottom of the tables indicates how many times a particular method outperforms others per metric per dataset.

For univariate setting, each method produces predictions for a single output over the time series. From Tab. \ref{tab:Model_accuracy}, we observe the following: 

\textbf{(1)} FWin achieves the best performance with total count of 32. 

\textbf{(2)} When comparing FWin and Informer, FWin outperforms the Informer by a margin of 50 to 16. 

\textbf{(3)} FWin performs well on the ETT, Exchange, ILI datasets, it remains competitive for the Weather dataset. However, it performs worst on the ECL dataset. This could be due to the differences caused by the dataset, which is common in time-series model \cite{li2022generalizable}, or Informer's ProbSparse hypothesis satisfied well for this dataset. 

\textbf{(4)} The average MSE reduction is 19.60\%, and MAE is about 11.88\%, when comparing  FWin with Informer.

For the multivariate setting, each method produces predictions based on multiple features over the time series. From Tab. \ref{tab:Model_accuracy}, we observe that: 

\textbf{(1)} The light model FWin-S leads the count at 34 total, followed closely by FWin with a total count of 30. 

\textbf{(2)} In a head to head comparison, FWin outperforms Informer by a large margin (59 to 7). 

\textbf{(3)} FWin and FWin-S have close accuracies, yet FWin is overall better on both uni/multi-variate data-sets. 

\textbf{(4)} The average MSE (MAE) reduction is about 16.33\% (10.96\%).

\begin{table*}[ht!]
\caption{Accuracy comparison on LSTF benchmark data (S/M denotes uni/multi-variate data), best results highlighted in bold.}
    \vskip 0.1in
    \centering
    \begin{tabular}{|c|c|c c|c c|c c|||c c|c c|c c|c c|c c|}
    \hline
         \multicolumn{2}{|c|}{Methods}& \multicolumn{2}{c|}{Informer (S)} & \multicolumn{2}{c|}{FWin (S)} & \multicolumn{2}{c|||}{FWin-S (S)} & \multicolumn{2}{c|}{Informer (M)} & \multicolumn{2}{c|}{FWin (M)} & \multicolumn{2}{c|}{FWin-S (M)} \\   
         \hline
         \multicolumn{2}{|c|}{Metric} & MSE & MAE & MSE & MAE & MSE & MAE & MSE & MAE & MSE & MAE & MSE & MAE\\
         \hline
         \multirow{5}{1em}{\rotatebox{90}{ETTh$_1$}} & 24 &0.116& 0.273 & \textbf{0.060}& \textbf{0.196}& 0.116& 0.272 &  0.528& 0.525& \textbf{0.483}& \textbf{0.499}& 0.507& 0.517\\
         & 48& 0.170& 0.333& \textbf{0.102}& \textbf{0.257}& 0.141& 0.302     &         0.764& 0.665& \textbf{0.638}& \textbf{0.592}& 0.695& 0.626 \\ 
         & 168& \textbf{0.149}& 0.311& 0.150& \textbf{0.308}& 0.252& 0.401 &         1.083& 0.836& 1.004& 0.786& \textbf{0.885}& \textbf{0.742}\\
         & 336& 0.160& 0.323& \textbf{0.108}& \textbf{0.261}& 0.397& 0.519&         1.270& 0.920& 1.094& 0.821& \textbf{1.022}& \textbf{0.814} \\
         & 720& 0.258& 0.428& \textbf{0.105}& \textbf{0.253}&  0.270&  0.434&         1.447& 0.977& 1.181 & 0.873& \textbf{1.087}& \textbf{0.846} \\
         \hline
         \hline
         \multirow{5}{1em}{\rotatebox{90}{ETTh$_2$}} & 24 &0.086& 0.225 & 0.082& 0.221& \textbf{0.075}& \textbf{0.212} & \textbf{0.455} & \textbf{0.508} &0.550& 0.566& 0.551& 0.568\\
         & 48& 0.164& 0.318&\textbf{0.125}& \textbf{0.277}& 0.140& 0.299&          2.368& 1.241& \textbf{0.774}& \textbf{0.664}& 0.792& 0.680 \\
         & 168& 0.270& 0.416& \textbf{0.220}& \textbf{0.375}& 0.277& 0.422  &          5.074& 1.910& \textbf{2.309}& \textbf{1.136}& 2.767& 1.327\\
         & 336& 0.324& 0.458& \textbf{0.244}& \textbf{0.396}& 0.277& 0.425 &          3.116& 1.460& \textbf{2.461}& \textbf{1.187}& 2.663& 1.337\\
         & 720& 0.294& 0.438& \textbf{0.261} & \textbf{0.412}& 0.266& 0.423 &          4.193& 1.778&  \textbf{2.847} & \textbf{1.286}& 3.258& 1.530\\
         \hline
         \hline
         \multirow{5}{1em}{\rotatebox{90}{ETTm$_1$}} & 24 &0.025& 0.122& \textbf{0.015}& \textbf{0.096}& 0.021& 0.112 & 0.346& 0.397& \textbf{0.305}& \textbf{0.375}& 0.322& 0.389\\
         & 48& 0.055& 0.181& \textbf{0.031}& \textbf{0.132}& 0.032& 0.135&          0.480& 0.482&  \textbf{0.408}& \textbf{0.444}& 0.436& 0.467 \\
         & 96& 0.181& 0.356& \textbf{0.050}& \textbf{0.175}& 0.086& 0.234 &          0.555& 0.531& \textbf{0.517}& \textbf{0.517}& 0.573& 0.553\\
         & 288& 0.279& 0.450& 0.179& 0.341& \textbf{0.173}& \textbf{0.334} &          0.943& 0.746& 0.831& \textbf{0.688}& \textbf{0.794}& 0.691\\
         & 672& 0.396& 0.559& \textbf{0.133} & \textbf{0.286}& 0.152&  0.314&         0.903& \textbf{0.729}& 1.119 & 0.838& \textbf{0.890}& 0.741\\ 
         \hline
         \hline
         \multirow{5}{1em}{\rotatebox{90}{Weather}} & 24 &0.113& 0.249 & \textbf{0.104}& \textbf{0.236}& 0.111& 0.243& 0.335& 0.388& \textbf{0.310}& \textbf{0.363}& 0.316& 0.369\\
         & 48& 0.196& 0.332& \textbf{0.172}& 0.315& 0.176& \textbf{0.310}& 0.395& 0.434&  \textbf{0.379}& 0.419& \textbf{0.379}& \textbf{0.415} \\
         & 168& 0.257& 0.376& 0.259& 0.391& \textbf{0.235}& \textbf{0.357}&         0.625& 0.580& 0.561& 0.539& \textbf{0.544}& \textbf{0.530}\\
         & 336& 0.275& 0.397& 0.338& 0.458& \textbf{0.262}& \textbf{0.388}&         0.665& 0.611& 0.630& 0.585& \textbf{0.598}& \textbf{0.574}\\
         & 720& 0.259& 0.389& 0.291 & 0.421& \textbf{0.250} & \textbf{0.383}&         0.657& 0.604& 0.686 & 0.614& \textbf{0.576}& \textbf{0.560}\\
         \hline
         \hline
         \multirow{5}{1em}{\rotatebox{90}{ECL}} & 48 &0.259& \textbf{0.359}& \textbf{0.249}& 0.369& 0.274& 0.388 & 0.293& 0.382&  0.291& \textbf{0.371}& \textbf{0.287}& 0.372\\
         & 168& \textbf{0.332}& \textbf{0.410}& 0.408& 0.479& 0.403& 0.472&         \textbf{0.292}& 0.386& 0.302& 0.379& 0.295& \textbf{0.378} \\
         & 336& \textbf{0.378}& \textbf{0.441}& 0.477& 0.516& 0.407& 0.471&          0.422& 0.467& 0.327& 0.399& \textbf{0.307}& \textbf{0.388}\\
         & 720& \textbf{0.373}& \textbf{0.444}& 0.568& 0.577& 0.376& 0.455 &          0.600& 0.559&  0.344& 0.408& \textbf{0.311}& \textbf{0.390} \\
         & 960& 0.365& 0.448& 0.415& 0.485& \textbf{0.359}& \textbf{0.448} &          0.871& 0.714&  0.362 & 0.421& \textbf{0.314}& \textbf{0.393} \\
         \hline
         \hline
         \multirow{4}{1em}{\rotatebox{90}{Exchange}} & 96 &0.305& 0.435& 0.294& \textbf{0.409}& \textbf{0.281}& 0.414 & 0.991& 0.797&  0.828& 0.733& \textbf{0.762}& \textbf{0.694}\\
         & 192& 1.345& 0.902& 0.679& 0.622& \textbf{0.566}& \textbf{0.574}& 1.175& 0.859& \textbf{1.148}& \textbf{0.889}& 1.178& \textbf{0.889} \\
         & 336& 2.441& 1.253& 0.949& 0.761& \textbf{0.785}& \textbf{0.703}& 1.581& 0.999& \textbf{1.301}& \textbf{0.960}& 1.344& 0.971\\
         & 720& 1.933& 1.106& \textbf{1.127}& \textbf{0.891}& 1.253& 0.897 & 2.643& 1.356&  2.071& 1.196& \textbf{2.036}& \textbf{1.177}\\
         \hline
         \hline
         \multirow{4}{1em}{\rotatebox{90}{ILI}} & 24 &5.404& 2.057& 3.727 & 1.648& \textbf{3.491}& \textbf{1.597}& 6.048& 1.698& 3.881& 1.308& \textbf{3.849}& \textbf{1.298}\\
         & 36& 4.384& 1.849& 3.124& 1.534& \textbf{3.023}& \textbf{1.528}& 5.871& 1.681& 4.036& 1.331& \textbf{4.024}& \textbf{1.316} \\
         & 48& 4.487& 1.886& 3.697& 1.680& \textbf{3.293}& \textbf{1.605}& 5.171& 1.551& \textbf{4.334}& \textbf{1.404}& 4.431& 1.406\\
         & 60& 5.179& 2.035& 4.141& 1.788& \textbf{3.767}& \textbf{1.734}& 5.273& 1.553& \textbf{4.547}& 1.443& 4.600& \textbf{1.438}\\
         \hline
         \hline
         \multicolumn{2}{|c|}{Count}& \multicolumn{2}{c|}{8}& \multicolumn{2}{c|}{32}& \multicolumn{2}{c|||}{26} &  \multicolumn{2}{c|}{4}& \multicolumn{2}{c|}{30}& \multicolumn{2}{c|}{34}  \\
         \hline
    \end{tabular}

\label{tab:Model_accuracy}
\vskip -0.1in
\end{table*}

\begin{table*}[ht!]
\caption{Post-fault voltage prediction accuracy comparison on power grid dataset (S/M same as in Tab. \ref{tab:Model_accuracy}), best results highlighted in bold.}
    \vskip 0.1in
    \centering
      \fontsize{8}{10}\selectfont
    \begin{tabular}{|c|c|c c|c c|c c|c c|c c|c c|c c|c c|c c|}
    \hline
         \multicolumn{2}{|c|}{Methods}& \multicolumn{2}{c|}{Informer} & \multicolumn{2}{c|}{FWin} & \multicolumn{2}{c|}{FWin-S} & \multicolumn{2}{c|}{FEDformer} & \multicolumn{2}{c|}{Autoformer} & \multicolumn{2}{c|}{ETSformer} & \multicolumn{2}{c|}{PatchTST}\\   
         \hline
         \multicolumn{2}{|c|}{Metric} & MSE & MAE & MSE & MAE & MSE & MAE & MSE & MAE & MSE & MAE & MSE & MAE & MSE & MAE\\
         \hline
           M& 700 &0.049& 0.113& 0.063 & 0.141& \textbf{0.043}& \textbf{0.101}& 0.245& 0.311& 0.390& 0.403& 0.339 & 0.381 & 0.211&0.278\\
           \hline
          S& 700 & 0.111 & 0.170& \textbf{0.091} & 0.141& 0.092 & \textbf{0.132} & 0.272 & 0.310 & 0.367& 0.388& 0.303 &0.322 & 0.138&0.188\\
         \hline
    \end{tabular}

\label{tab:power_accuracy}
\vskip -0.1in
\end{table*}

\begin{figure}[ht!]
    \centering
    \includegraphics[width=\columnwidth]{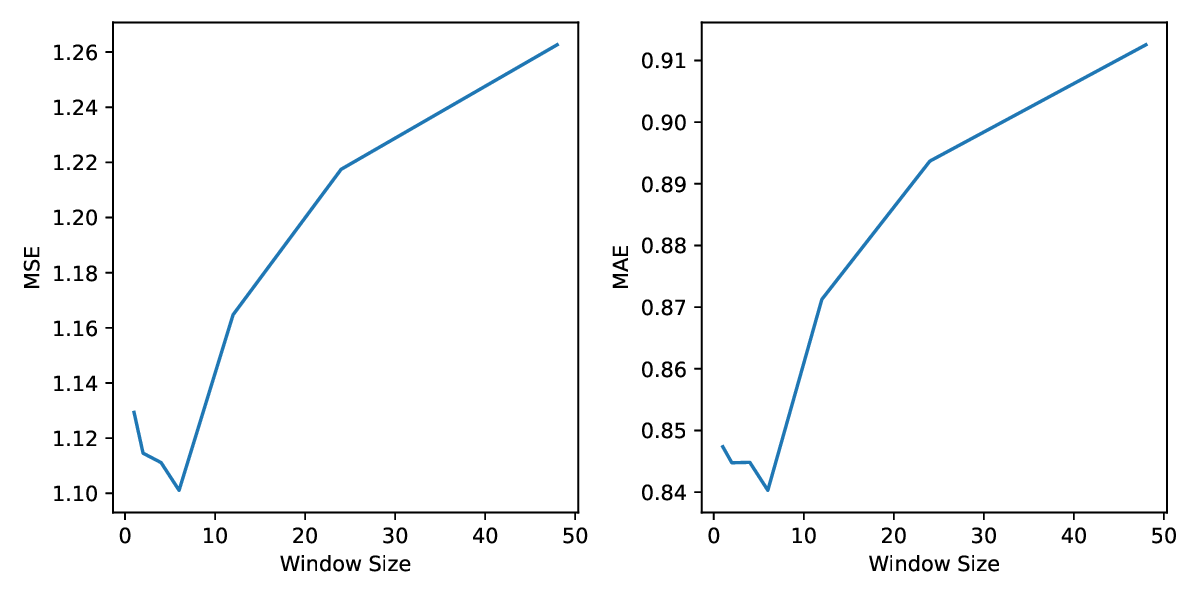}
    \caption{Window size versus error for ETTh1 multivariate data on the long range prediction metric of 720.}
    \label{fig:windowsize_vs_error}
\end{figure}

Optimizing the window size for each data set may increase performance of our model. In Fig. \ref{fig:windowsize_vs_error}, we present an ablation study of the effect of window sizes to the error for ETTh1 on the long range metric of 720. In this case, a smaller window size, i.e. 6, would provide the best result. However, to keep the experiments consistent, we decide to keep the window size a fixed constant of 24. 
The time scale of a dataset may impact the choice of window size. 
Many of the datasets have hourly time scale, thus choosing a window size of 24 is meaningful in covering a daily observation. Refer to Appendix \ref{sec:effect_window} for full study of the effect of window sizes.

The experiments in the tables were conducted using similar hyper-parameters to those of Informer \cite{informer_21} and Autoformer \cite{autoformer_21}. Interestingly, FWin-S does so well without the Fourier Mix block in its decoder. 

\begin{figure*}[ht!]
    \centering
    \includegraphics[width=0.9\textwidth]{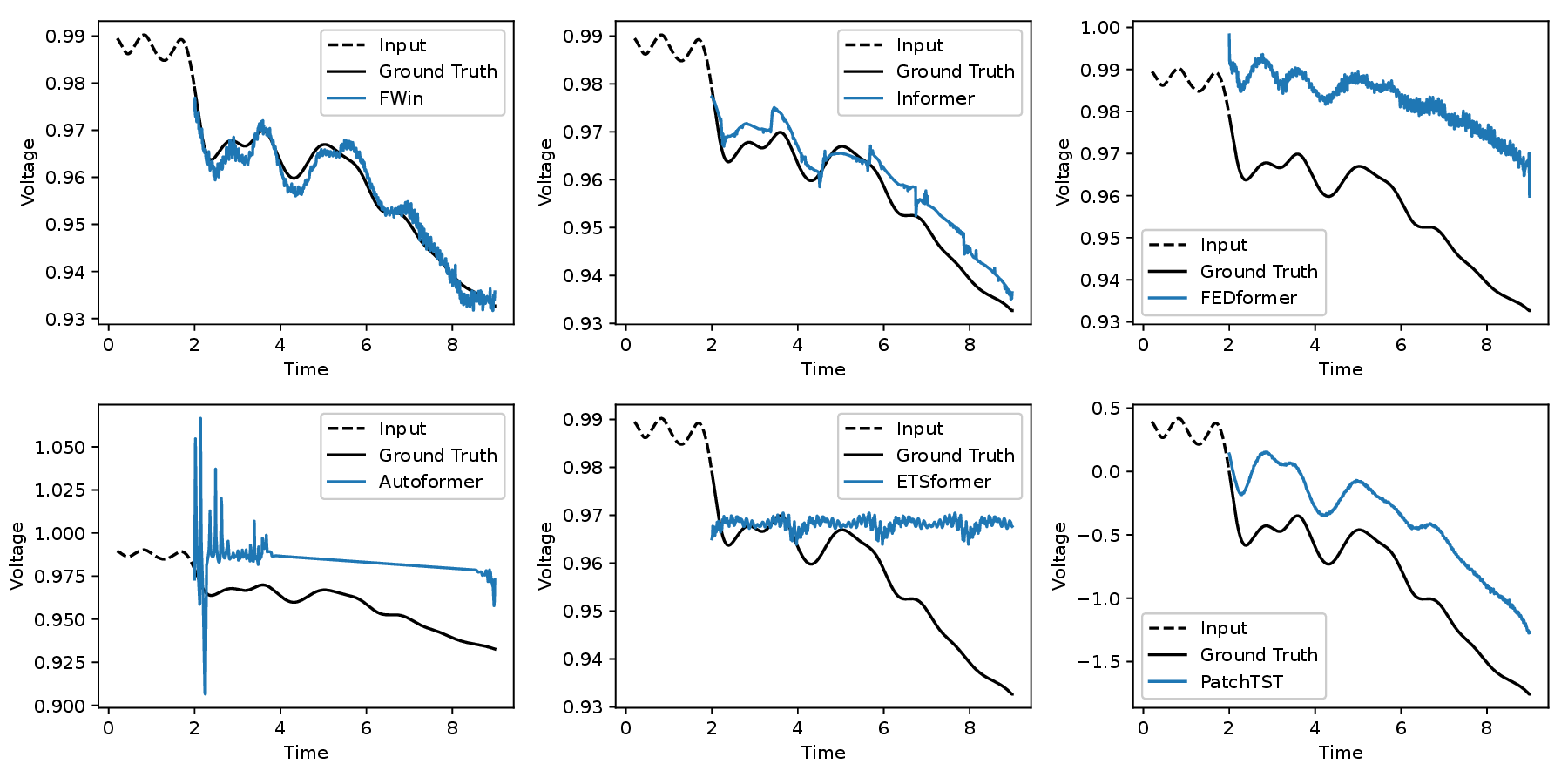}
    \caption{Univariate post fault prediction comparison 
    (voltage vs. time in second) on power grid data \cite{powergrid_toolbox,glassoformer_22}: \{FWin,Informer\} outperform (FED,Auto,ETS)formers and PatchTST. The dashed line under 2 second duration is the input, to the right of which are the predictions vs. the ground truth (in black).}
    \label{fig:power_grid_prediction}
\end{figure*}

Informer's prediction accuracies on the benchmark datasets in Tab. \ref{tab:Model_accuracy} have been largely improved by recent transformers such as PatchTST \cite{patchtstnie2023}, Autoformer \cite{autoformer_21}, FEDformer \cite{fedformer} and ETSformer \cite{woo2022etsformer} designed  with certain prior-knowledge of datasets,  e.g. using auto-correlation or trend/seasonality decomposition. Like Informer, FWin has no prior-knowledge based 
operation, which helps to generalize better on non-stationary time series where seasonality is absent. 
Such a situation arises in post-fault decision making on a power grid where predicting 
transient trajectories is important for system operators to take appropriate actions \cite{MLGDyn}, e.g., a load shedding upon a voltage
or frequency violation.
We carry out experiments 
on a simulated New York/New England 16-generator 68-bus power system \cite{powergrid_toolbox,glassoformer_22}. The system has 88 lines linking the buses, and can be regarded as a graph with 68 nodes and 88 edges. 
The data set has over 2000 fault events, where each event has signals of 10 second duration. These signals contain voltage and frequency from every bus, and current from every line. The train/val/test split is 1000:350:750.
Tab. \ref{tab:power_accuracy} shows that FWin and FWin-S improve or maintain Informer's 
accuracy in a robust fashion while the four
recent transformers pale considerably in comparison.
Fig. \ref{fig:power_grid_prediction} illustrates model predictions on the power grid dataset \cite{powergrid_toolbox,glassoformer_22}. FWin and Informer outperform FEDformer, Autoformer, ETSformer, and PatchTST. 

Besides performance,  training and inference times of the models are our main concern and are summarized below: 

\textbf{(1)} Compared to Informer, FWin achieves average speed up factors of \textbf{1.7} and \textbf{1.4} for inference and training times respectively. The ILI dataset has the lowest speed-up factor because both the input and the prediction lengths are small. See Tab. \ref{tab:Avg_SpeedUp_Train_Inference} and Appendix C for breakdowns on six benchmark datasets, {\it similarly on other datasets in this paper}.

\textbf{(2)} FWin inference and training  times are very close to those of FWin-S. This indicates that the Fourier Mix layer in the decoder adds a minimal overhead to the overall model.
The FWin-S model exhibits the fastest inference and training times, as expected because it is the model with smallest parameter size here.
 
\textbf{(3)} FWin has approximately $8.1$ million parameters, whereas Informer has around $11.3$ million parameters under default settings, resulting in a reduction about $28\%$.

 

\textbf{(4)} The inference time for Informer increases with prediction length. FWin's inference time exhibits minimal growth. The Exchange dataset demonstrates this effect as we used the same input length for all prediction lengths, and ran the models on a single GPU (see Appendix \ref{subsec:fwin_vs_informer_time}).


\begin{table}[ht!]
    \caption{Average inference/training speed-up factors of FWin vs. Informer,  see Appendix C for machine times.}  
    \vskip 0.1in
    \centering
    \begin{tabular}{|c|c|c|c|}
        \hline
         Data& Feature& Inference & Train\\
         \hline
         ETTh$_1$& Multivariate & 1.66& 1.34\\
         ETTh$_1$& Univariate & 1.80& 1.64\\
         \hline
         \hline
         ETTm$_1$& Multivariate & 1.70& 1.30\\
         ETTm$_1$& Univariate & 1.68& 1.34\\
         \hline
         \hline
         Weather& Multivariate & 1.70 & 1.28 \\
         Weather& Univariate & 1.99 & 2.01 \\
         \hline
         \hline
         ECL& Multivariate & 1.59& 1.10\\
         ECL& Univariate & 2.01& 1.53\\
         \hline
         \hline
         Exchange& Multivariate & 1.77& 1.25\\
         Exchange& Univariate & 1.69& 1.25\\
         \hline
         \hline
         ILI& Multivariate & 1.56& 1.19\\
         ILI& Univariate & 1.62& 1.29\\
         \hline
         
    \end{tabular}
   
    \label{tab:Avg_SpeedUp_Train_Inference}
    \vskip -0.1in
\end{table}

\section{Theoretical Results}


\begin{definition}\label{def:bdi_matrix}
    Let $A\in\mathbb{R}^{L\times L}$ with $a_{ij}$ its $(i,j)$-th entry. Let $w\in\mathbb{N}$ be a factor of 
    $L$; and for 
    $\forall n\in \{1,\dots, L/w\}$, let $A_n$ be the sub-matrix of $A$ with entries  
    $(a_{ij})_{i=w(n-1)+1,j=w(n-1)+1}^{i=nw, j=nw}$. We say $A$ is block diagonally invertible ({\bf BDI}) if for every $n$, $A_n$ is invertible.
\end{definition}


\begin{definition}\label{def:attn_matrix}
    Let $Q, K\in\mathbb{R}^{L\times d}$ be the query, key matrix respectively. Define the attention matrix as:
\begin{equation}\label{eq:attn_matrix}
    Attn(Q,K):= \text{softmax}(QK^T/\sqrt{d}).
\end{equation}
\end{definition}

\begin{definition}\label{def:full_attn}
    Let $Q, K, V\in\mathbb{R}^{L\times d}$ be the query, key, value matrix respectively. Define the full attention as:
\begin{equation}\label{eq:full_attn}
    Attn_f(Q,K,V):= Attn(Q,K)V.
\end{equation}
\end{definition}

\begin{definition}\label{def:win_attn}
    Let $Q, K, V\in\mathbb{R}^{L\times d}$ be the query, key, value matrix respectively with $q_i, k_i, v_i$ the $i$-th row of the matrix $Q, K, V$. Let $w\in\mathbb{N}$ be the window size, such that $w$ divides $L$.  Define the window attention as:
    \begin{equation}\label{eq:win_attn}
        Attn_w(Q, K, V, w):= 
        \begin{bmatrix}
             \sum_{j\in J(1)} \frac{\exp(q_1k_j^T)v_j}{\gamma_1}\\
            \vdots\\
            \sum_{j\in J(L)} \frac{\exp(q_Lk_j^T)v_j}{\gamma_L}
        \end{bmatrix}.
    \end{equation}
    Here $J(m) = \{Mw +1,\dots, (M+1)w\}$, where $M = \lfloor\frac{m-1}{w}\rfloor$; and \begin{equation}
    \gamma_m = \sum_{j\in J(m)} \exp(q_mk_j^T/\sqrt{d}).
\end{equation}

\end{definition}

\begin{definition}\label{def:mix_win_attn_matrix}
    Let $A\in\mathbb{R}^{L\times L}$ and $Q, K, V, w$ be the same as in \textbf{Definition \ref{def:win_attn}}, define the mixed window attention as:
\begin{equation}\label{eq:mix_win_attn_matrix}
    Attn_{mw}(Q,K,V,w, A):= A\,  Attn_w(Q,K,V,w)
\end{equation}
\end{definition}

\begin{theorem}\label{thm:full_attn=mw_attn}
    Let $Q, K, V \in\mathbb{R}^{L\times d}$. Let $w\in \mathbb{N}$ such that $w$ divides $L$. If Attn($Q,K$) is BDI, then there exists a matrix $A\in \mathbb{R}^{L\times L}$ such that 
    \begin{equation}
        Attn_f(Q,K,V) = \text{Attn}_{mw}(Q,K,V,w, A).
    \end{equation}
    In particular, we can construct the exact value of $A$.
\end{theorem}

We provide the details of the proof in Appendix \ref{sec:theo_results_proof}.


\begin{definition}\label{def:fwin_attn}
    Let $A\in\mathbb{R}^{L\times L}$ and $Q, K, V, w$ be the same as defined in \textbf{Definition \ref{def:win_attn}}, define the Fourier-mixed window attention as:
\begin{equation}\label{eq:fwin_attn_matrix}
    Attn_{Fwin}(Q,K,V,w, A):= A \, \mathcal{F}(Attn_w(Q,K,V,w)),
\end{equation}
where $\mathcal{F}$ is the discrete Fourier transform.
\end{definition}

\begin{corollary}\label{cor:fwin_attn=full_attn}
    Let $Q, K, V$ and $w$ be the same as defined in \textbf{Theorem \ref{thm:full_attn=mw_attn}}. If Attn($Q,K$) is BDI, then there exists a matrix $A\in \mathbb{C}^{L\times L}$ such that 
    \begin{equation}
        \text{Attn}_f(Q,K,V) = \text{Attn}_{Fwin}(Q,K,V,w, A).
    \end{equation}
\end{corollary}

\begin{definition}\label{def:hwin_attn}
    Let $A\in\mathbb{R}^{L\times L}$ and $Q, K, V, w$ be the same as defined in \textbf{Definition \ref{def:win_attn}}, define the Hartley-mixed window attention as:
\begin{equation}\label{eq:hwin_attn_matrix}
    Attn_{Hwin}(Q,K,V,w, A):= A \, \mathcal{H}(Attn_w(Q,K,V,w)),
\end{equation}
where $\mathcal{H}$ is the Hartley transform \cite{bracewell1986hartley}.
\end{definition}

\begin{corollary}\label{cor:hwin_attn=full_attn}
    Let $Q, K, V$ and $w$ be the same as defined in \textbf{Theorem \ref{thm:full_attn=mw_attn}}. If Attn($Q,K$) is BDI, then there exists a matrix $A\in \mathbb{R}^{L\times L}$ such that 
    \begin{equation}
        \text{Attn}_f(Q,K,V) = \text{Attn}_{Hwin}(Q,K,V,w, A).
    \end{equation}
\end{corollary}



\section{Conclusion}

We introduced FWin Transformer and its light weight version FWin-S to successfully accelerate Informer by replacing its ProbSparse and full attention layers with window attention and Fourier mixing blocks in both encoder and decoder. The FWin attention approach does not rely on sparse attention hypothesis or periodic like patterns in the data.
The experiments on univariate and multivariate datasets and theoretical guarantee demonstrated FWin's merit 
in fast inference on long sequence time-series forecasting while improving or maintaining Informer's performance. 

In future work, we plan to  
%
further optimize and extend the FWin approach for 
accelerating transformers (including the prior-knowledge based ones) in 
various 
applications.

\newpage
\appendix
\section{Related Work}

\begin{table*}[ht!]
    \caption{Inference/training time comparison in seconds on LSTF benchmark data (S/M denotes uni/multivariate data, Inf denotes inference time).}
    \vskip 0.1in
    \centering
     \begin{threeparttable}[b]
    \begin{tabular}{|c|c|cc|cc|cc||cc|cc|cc|}
    \hline
         \multicolumn{2}{|c|}{Methods}& \multicolumn{2}{c|}{Informer (S)} & \multicolumn{2}{c|}{FWin (S)} & \multicolumn{2}{c||}{FWin-S (S)}& \multicolumn{2}{c|}{Informer (M)} & \multicolumn{2}{c|}{FWin (M)} & \multicolumn{2}{c|}{FWinS (M)} \\
         \hline
         \multicolumn{2}{|c|}{Metric} & Inf & Train & Inf & Train & Inf & Train& Inf & Train & Inf & Train & Inf & Train\\
         \hline
         \multirow{5}{1em}{\rotatebox{90}{ETTh$_1$}} 
         & 24 & 0.280\tnote{*}& 803.48\tnote{*} & 0.112 & 267.6& 0.107 & 262.0& 0.153 & 75.9 & 0.100 & 50.8 & 0.094& 49.1\\
         & 48 & 0.269& 687.4& 0.172 & 528.9& 0.168 & 512.3 & 0.156 & 101.7 & 0.102 & 73.3 & 0.099 & 71.1 \\
         & 168 & 0.297& 718.4& 0.180 & 548.3& 0.176 & 537.9 & 0.175 & 205.5 & 0.106 & 156.4 & 0.101 & 147.7 \\
         & 336 & 0.279& 706.7& 0.179 & 545.0& 0.165 & 534.6& 0.207 & 253.5 & 0.110 & 195.8 & 0.106 & 184.2 \\
         & 720 & 0.303& 671.4& 0.176 & 516.0& 0.171 & 504.9& 0.321 & 676.8 & 0.179 & 557.4& 0.176 & 554.7\\
         \hline
         \hline
         \multirow{5}{1em}{\rotatebox{90}{ETTm$_1$}} 
         & 24 & 0.153& 381.4& 0.102 & 281.3& 0.098 & 263.5& 0.193& 1567.2& 0.108 & 1029.7& 0.102 & 1000.5 \\
         & 48 & 0.155& 411.7& 0.099 & 310.6& 0.096 & 292.4 & 0.158& 417.8& 0.116 & 317.7& 0.095 & 296.4  \\
         & 168 & 0.198& 1476.0& 0.108 & 1065.4& 0.104 & 1027.3& 0.200& 1501.6& 0.112 & 1102.6& 0.104 & 1047.3 \\
         & 288 & 0.278& 2882.8& 0.171 & 2203.9& 0.166 & 2161.5& 0.312& 3046.3& 0.184 & 2690.9& 0.165 & 2304.4 \\
         & 672 & 0.336& 3058.9& 0.178 & 2284.1& 0.176 & 2259.0& 0.368& 3206.6& 0.196 & 2761.3& 0.176& 2484.3 \\
         \hline
         \hline
         \multirow{5}{1em}{\rotatebox{90}{Weather}} 
         & 24 & 0.280\tnote{*}& 2230.9\tnote{*} & 0.111 & 734.8& 0.106 & 729.6& 0.228 & 623.4 & 0.142 & 474.9 & 0.136& 453.9\\
         & 48 & 0.277\tnote{*}& 2468.9\tnote{*}& 0.114 & 753.7& 0.110 & 741.8& 0.170 & 318.2 & 0.101 & 237.0 & 0.097 & 229.9 \\
         & 168 & 0.233& 844.3& 0.153 & 676.2& 0.144 & 632.5 & 0.257 & 1084.2 & 0.153 & 819.1 & 0.151 & 790.7 \\
         & 336 & 0.389& 2558.2& 0.241 & 2214.5& 0.236 & 2098.3 & 0.419 & 2668.3 & 0.248 & 2257.2 & 0.229 & 2214.0 \\
         & 720 & 0.476& 2756.2& 0.257 & 2399.0& 0.243 & 2193.8 & 0.554 & 2911.3 & 0.283 & 2380.6 & 0.279 & 2331.9\\
         \hline
         \multirow{5}{1em}{\rotatebox{90}{ECL}} 
         & 48 & 0.276\tnote{*}& 1354.2\tnote{*}& 0.109 & 558.9& 0.107& 554.5& 0.165 & 324.3 & 0.125 & 290.8 & 0.100 & 243.3 \\
         & 168 & 0.218& 925.8& 0.117 & 626.3& 0.113& 618.1 & 0.202 & 811.1& 0.121 & 696.6& 0.112 & 624.0 \\
         & 336 & 0.217& 785.7& 0.114 & 587.1& 0.111& 565.0 & 0.225 & 1462.3& 0.138& 1297.5& 0.130& 1167.9\\
         & 720 & 0.315& 1337.8& 0.180 & 1128.3& 0.175& 1117.7& 0.371 & 3400.7& 0.227& 3352.1& 0.220& 3101.7 \\
         & 960 & 0.417& 1445.3& 0.205 & 1171.8& 0.199& 1121.4 & 0.415& 3703.0& 0.246& 3445.8&  0.235& 3398.6 \\
         \hline
         \multirow{4}{1em}{\rotatebox{90}{Exchange}} 
         & 96 & 0.161& 65.2& 0.099 & 49.3& 0.097& 47.9& 0.158 & 65.7 & 0.100 & 51.0 & 0.098 & 49.6 \\
         & 192 & 0.164& 79.7& 0.102 & 62.6& 0.099& 59.9 & 0.164 & 81.9& 0.103 & 64.4& 0.100 & 61.5 \\
         & 336 & 0.178& 98.0& 0.111 & 80.1& 0.103& 76.2 & 0.177 & 101.2& 0.108& 82.0& 0.102& 79.1\\
         & 720 & 0.226& 145.4& 0.118 & 121.2& 0.115& 114.7& 0.271 & 149.1& 0.120& 124.1& 0.114& 117.8 \\
         \hline
         \multirow{4}{1em}{\rotatebox{90}{ILI}} 
         & 24 & 0.155& 7.9& 0.099 & 5.8& 0.097& 5.5& 0.152 & 9.4 & 0.097 & 7.0 & 0.094 & 6.9 \\
         & 36 & 0.153& 7.8& 0.100 & 5.8& 0.096& 5.5 & 0.156 & 9.0& 0.099 & 8.7& 0.093 & 7.0 \\
         & 48 & 0.150& 7.4& 0.097 & 6.0& 0.097& 5.7 & 0.155 & 9.1& 0.100& 7.3& 0.096& 7.3\\
         & 60 & 0.181& 9.3& 0.099 & 7.6& 0.097& 6.7& 0.152 & 9.0& 0.099& 8.0& 0.096& 7.3 \\
         \hline
    \end{tabular}
    \begin{tablenotes}
        \item [*] using multiple GPUs compared to results by FWin/FWin-S on single GPU.
    \end{tablenotes}

    \label{tab:Train_Inference_Time}
    \end{threeparttable}
    \vskip -0.1in
    \end{table*}

Several types of attention models are related to our work here.

First, MLP mixers relax the graph similarity constraints of the self-attention and mix tokens with MLP projections. The original MLP-mixer \cite{mlpmixer_21} reaches similar
accuracy as self-attention in vision tasks. However, such a method lacks scalability as a result of quadratic complexity of MLP projection, and suffers from parameter inefficiency for high resolution input.

Next, Fourier-based mixers apply Fourier transform to mix tokens in NLP and vision tasks. FNet \cite{Fnet}
 resembles the MLP-mixer with token mixer being the classical discrete Fourier transform (DFT), without adaptive filtering on data distribution. Global filter networks (GFNs \cite{GFN_21}) learn filters in the Fourier domain to perform depth-wise global convolution with no channel mixing involved. Also,
GFN filters lack adaptivity that could negatively impact generalization. 
AFNO \cite{adafno_22} performs global convolution with dynamic filtering and channel mixing for better expressiveness and generalization. However, AFNO  network parameter sizes tend to be much larger than those of the light weight vision transformer (ViT) models such as GFN-T  \cite{GFN_21}, shift-window mixer Swin-T \cite{Swin} and hybrid convolution-attention models MOAT-T \cite{moat_22}, and mobile ViT \cite{mobile_vit_22}.

\begin{figure}[ht!]
    \centering
    \includegraphics[width=\columnwidth]{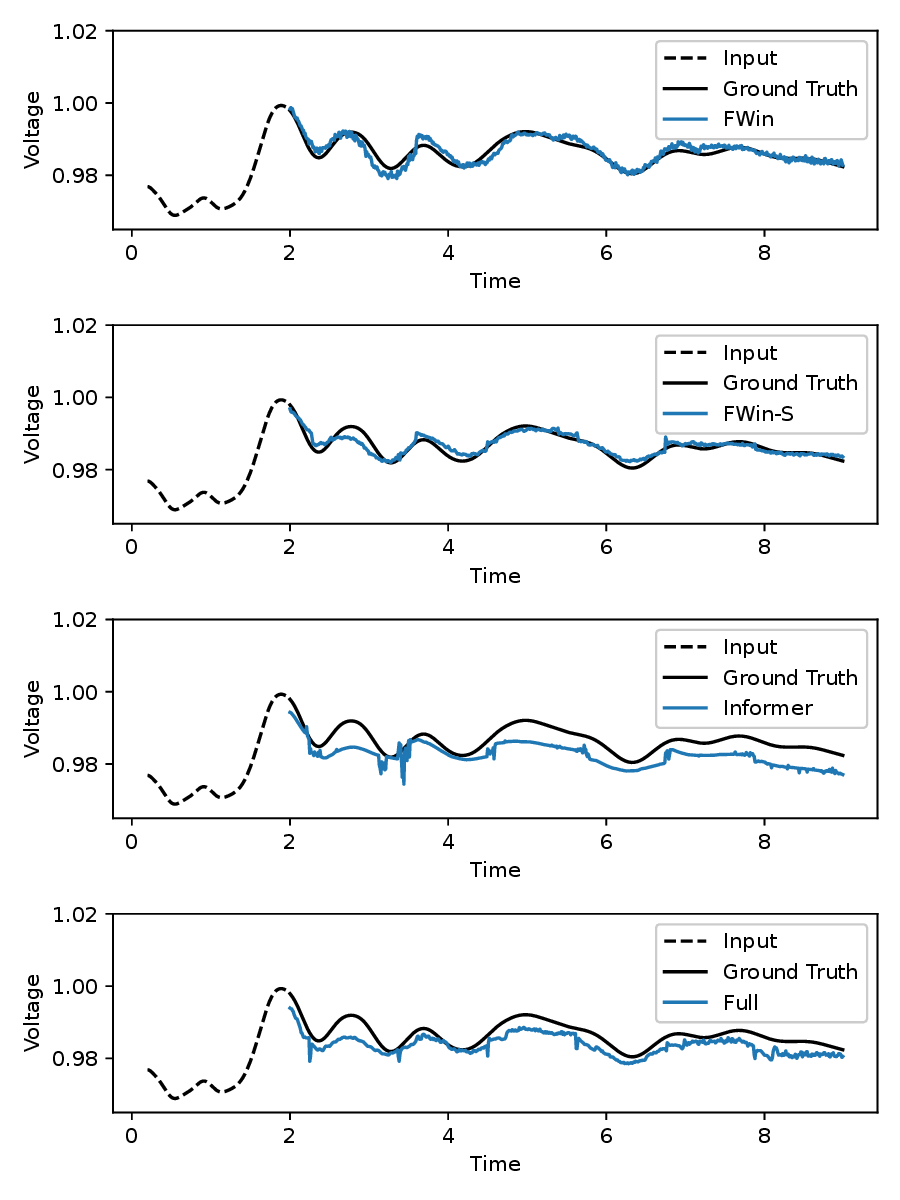}
    \caption{Univariate post fault prediction (voltage vs. time in second) on power grid data \cite{powergrid_toolbox,glassoformer_22}. FWin, FWin-S have ``smooth" predictions while Informer has spurious jumps. Full in the bottom frame refers to Informer using full attention instead of probsparse. The dashed line to the left of 2 second is the input, to the right of which are the model predictions vs. the ground truth (in black).}
    \label{fig:power_grid_prediction_spikes_b517}
\end{figure}

For long sequence time series forecasting, the Informer \cite{informer_21, haoyietal-informerEx-2023} has a hybrid convolution-attention design with a probabilistic sparsity promoting function (ProbSparse) to reduce complexity of the standard self-attention and cross-attention. 
We shall adopt Informer as our baseline in this work, as it compares favorably vs. efficient transformers in recent 
years (see Tab. 1 and Tab. 2 in \cite{informer_21}). More recent improvements on benchmark data sets include
 Autoformer \cite{autoformer_21}, FEDformer \cite{fedformer} and ETSformer \cite{woo2022etsformer}, which are designed with certain prior-knowledge of datasets,  e.g. using trend/seasonality decomposition and auto-correlation functions. 
 However, they are not as robust on non-stationary time series as Informer (Tab. 5). Informer's prediction is seen to generate spurious peaks on power grid data (3rd frame of Fig. \ref{fig:power_grid_prediction_spikes_b517}), while FWin predictions appear  smoother and free from such distortions. 
 Comparing Informer with full Informer in the bottom frame of Fig. \ref{fig:power_grid_prediction_spikes_b517}, we see that the cause of these distortions may be attributed to probsparse. Glassoformer \cite{glassoformer_22} uses group lasso penalty to enforce query sparsity and reduce complexity of full attention to speed up inference. Though this method works well on power grid data, its training time is higher than Informer since full attention is involved in network training.

\begin{table}[ht]
    \caption{FWin vs. FNet \cite{Fnet} (replacing ProbSparse attention of Informer by Fourier-Mix followed by an FC layer) on multivariate ETTh1 data.}
    \vskip 0.1in
    \centering
    \begin{tabular}{|c|c c|c c|}
    \hline
         Model& \multicolumn{2}{c|}{FNet} & \multicolumn{2}{c|}{FWin} \\
    \hline
         Metric& MSE & MAE & MSE & MAE \\ 
    \hline
         24 & 0.490& 0.502& \textbf{0.483} & \textbf{0.499}\\
         48 & \textbf{0.562}& \textbf{0.543}& 0.638& 0.592\\
         168 & 1.052& 0.806& \textbf{1.004}&\textbf{0.786}\\
         336 & 1.194& 0.869& \textbf{1.094}&\textbf{0.821}\\
         720 & 1.349& 0.948& \textbf{1.181}&\textbf{0.873}\\
    \hline
     \multicolumn{1}{|c|}{Count}& \multicolumn{2}{c|}{2}& \multicolumn{2}{c|}{8} \\
     \hline
    \end{tabular}

    \label{tab:AbalationFWinvsFNet}
    \vskip -0.1in
\end{table}

\begin{table}[ht]
    \caption{FWin vs. Swin \cite{Swin} (replacing Fourier Mix in FWin by a shifted window attention) on multivariate ETTh1 data.}
    \vskip 0.1in
    \centering
    \begin{tabular}{|c|c c|c c|}
    \hline
         Model& \multicolumn{2}{c|}{Swin} & \multicolumn{2}{c|}{FWin} \\
    \hline
         Metric& MSE & MAE & MSE & MAE \\ 
    \hline
         24 & 0.567& 0.540& \textbf{0.483}& \textbf{0.499}\\
         48 & 0.685& 0.627& \textbf{0.638}& \textbf{0.592}\\
         168 & 1.016& 0.810& \textbf{1.004}& \textbf{0.786}\\
         336 & 1.161& 0.877& \textbf{1.094}&\textbf{0.821}\\
         720 & \textbf{1.095}& \textbf{0.844}& 1.181& 0.873\\
    \hline
     \multicolumn{1}{|c|}{Count}& \multicolumn{2}{c|}{2}& \multicolumn{2}{c|}{8} \\
     \hline
    \end{tabular}

    \label{tab:AbalationFWinvsWin}
    \vskip -0.1in
\end{table}

\section{Ablation Study}

\begin{table*}[ht!]
\caption{Model accuracy comparison on traffic forecasting (S/M denotes univariate/multivariate data), best results highlighted in bold. The relative MSE difference between Informer and FWin on S/M is 
5\% /12.23\%. 
The relative MAE difference between Informer and FWin on S/M data is 5.89\% /11.68\%.}
    \vskip 0.1in
    \centering
    \begin{tabular}{|c|c|c c|c c|c c|||c c|c c|c c|c c|c c|}
    \hline
         \multicolumn{2}{|c|}{Methods}& \multicolumn{2}{c|}{Informer (S)} & \multicolumn{2}{c|}{FWin (S)} & \multicolumn{2}{c|||}{FWin-S (S)} & \multicolumn{2}{c|}{Informer (M)} & \multicolumn{2}{c|}{FWin (M)} & \multicolumn{2}{c|}{FWinS (M)} \\   
         \hline
         \multicolumn{2}{|c|}{Metric} & MSE & MAE & MSE & MAE & MSE & MAE & MSE & MAE & MSE & MAE & MSE & MAE\\
         \hline
          \multirow{4}{1em}{\rotatebox{90}{Traffic}} & 96 &\textbf{0.192}& \textbf{0.284}& 0.199 & 0.294& 0.238& 0.315& 0.722& 0.395& \textbf{0.660}& \textbf{0.367}& 0.676& 0.375\\
         & 192& \textbf{0.212}& \textbf{0.301}& 0.218& 0.315& 0.245& 0.322& 0.779& 0.422& \textbf{0.672}& \textbf{0.368}& 0.693& 0.382 \\
         & 336& \textbf{0.223}& \textbf{0.310}& \textbf{0.223}& 0.321& 0.246& 0.328& 0.771& 0.423& \textbf{0.699}& \textbf{0.383}& 0.715& 0.392\\
         & 720& \textbf{0.246}& \textbf{0.334}& 0.266& 0.364& 0.340& 0.410& 0.892& 0.484& 0.761& 0.414& \textbf{0.704}& \textbf{0.383}\\
         \hline
    \end{tabular}

\label{tab:traffic_accuracy}
\vskip -0.1in
\end{table*}
We examined the benefits of combining window attention and Fourier mixing by
experiments on the ETTh1 dataset. Tab. \ref{tab:AbalationFWinvsFNet} shows that Fourier-mixed window attention outperforms using Fourier mixing alone (FNet \cite{Fnet}) in accuracy. 
 Tab.  \ref{tab:AbalationFWinvsWin} suggests that Fourier-mixed window attention is better overall than shifted window attentions in the encoder.  In view of Tab. \ref{tab:Model_accuracy}, FWinS is better than Swin on metric 720 in MSE (comparable in MAE).

\section{Effect of window size parameter}\label{sec:effect_window}

\begin{table*}[ht!]
\caption{Accuracy comparison of FWin model using different window sizes for various dataset on the multivariate task. Best results highlighted in bold.}
    \centering
    \begin{tabular}{|c|c|cc|cc|cc|cc|cc|cc|cc|cc|cc|}
    \hline
         \multicolumn{2}{|c|}{Window Size}&\multicolumn{2}{c|}{1}& \multicolumn{2}{c|}{2}& \multicolumn{2}{c|}{4} & \multicolumn{2}{c|}{6} & \multicolumn{2}{c|}{12} & \multicolumn{2}{c|}{24} \\   
         \hline
         \multicolumn{2}{|c|}{Metric} & MSE & MAE & MSE & MAE & MSE & MAE & MSE & MAE  & MSE & MAE & MSE & MAE\\
         \hline
         \hline
         \multirow{4}{1em}{\rotatebox{90}{ETTh1}} 
         & 24& 0.468& 0.486& 0.474& 0.490& 0.481& 0.498& \textbf{0.464}& \textbf{0.484}& 0.489& 0.505& 0.506& 0.515 \\
         & 48& 0.599& 0.565& 0.606& 0.571& 0.611& 0.580& 0.592& 0.569& \textbf{0.575}& \textbf{0.554}& 0.586& 0.561 \\
         & 168& 0.915& 0.741& 0.930& 0.748& 0.917& 0.752& \textbf{0.903}& \textbf{0.737}& 0.904& 0.741& 1.076& 0.822 \\
         & 336& 1.049& 0.782& 1.002& 0.769& 1.000& 0.774& \textbf{0.978}& \textbf{0.766}& 0.998& 0.772& 1.070& 0.813 \\
         & 720& 1.096& 0.837& 1.090& 0.840& \textbf{1.082}& \textbf{0.834}& 1.107& 0.844& 1.133& 0.859& 1.205& 0.884 \\
         \hline
         \hline
         \multirow{4}{1em}{\rotatebox{90}{Weather} } 
         & 24& 0.310& 0.361& 0.310& 0.365& 0.311& 0.364& \textbf{0.308}& \textbf{0.361}& \textbf{0.308}& 0.362& 0.309& 0.362 \\
         & 48& 0.382& \textbf{0.419}& 0.380& 0.421& \textbf{0.378}& 0.420& 0.380& 0.420& 0.381& 0.421& 0.381& 0.421 \\
         & 168& 0.561& 0.542& 0.561& 0.542& 0.570& 0.547& \textbf{0.558}& 0.539& 0.550& \textbf{0.535}& 0.561& 0.539 \\
         & 336& 0.631& 0.592& 0.626& 0.586& \textbf{0.610}& \textbf{0.577}& 0.629& 0.587& 0.612& 0.578& 0.618& 0.580\\
         \hline
         \hline
         \multirow{4}{1em}{\rotatebox{90}{Exchange}} 
         & 96 & 0.738& 0.695& 0.798& 0.723& 0.793& 0.717& \textbf{0.714}& \textbf{0.683}& 0.775& 0.717& 0.806& 0.729 \\
         & 192& 1.195& 0.910& \textbf{1.066}& \textbf{0.869}& 1.297& 0.951& 1.261& 0.941& 1.132& 0.891& 1.136& 0.886 \\
         & 336& 1.314& 0.964& \textbf{1.244}& 0.941& 1.340& 0.970& 1.270& \textbf{0.940}& 1.389& 0.992& 1.302& 0.962 \\
         & 720& 1.916& 1.134& 2.034& 1.166& 1.944& 1.137& \textbf{1.885}& \textbf{1.128}& 2.225& 1.225& 2.055& 1.176 \\

         \hline
         \hline
         \multicolumn{2}{|c|}{Count}& \multicolumn{2}{c|}{1}& \multicolumn{2}{c|}{3}& \multicolumn{2}{c|}{5}& \multicolumn{2}{c|}{14} &  \multicolumn{2}{c|}{4}&  \multicolumn{2}{c|}{0} \\
         \hline
    \end{tabular}

\label{tab:windowsize_vs_metric}
\end{table*}

Window size is an important parameter for FWin. In this section we will explore the effect of window size to model performance across many datasets presented in the paper. We present the results in Tab. \ref{tab:windowsize_vs_metric}. We observe that across the datasets, window size of 6 provides the best results overall. Window size of 1 provides competitive results compare to the best window size of 6. In general, under various window sizes, the performance is consistent.

\section{Theoretical Results}\label{sec:theo_results_proof}

\begin{definition}\label{def:bdi_matrix_proof}
    Let $A\in\mathbb{R}^{L\times L}$, with the $(i,j)$-th  entry of $A$ denoted by $a_{ij}$. Let $w\in\mathbb{N}$ be a factor of $L$. For every $n\in \{1,\dots, L/w\}$, let $A_n$ be the sub-matrix of $A$ such that the entries of $A_n$ are composed of $(a_{ij})_{i=w(n-1)+1,j=w(n-1)+1}^{i=nw, j=nw}$. We say $A$ is block diagonally invertible (BDI) if for every $n$, $A_n$ is invertible.
\end{definition}

\begin{definition}\label{def:attn_matrix_proof}
    Let $Q, K\in\mathbb{R}^{L\times d}$ be the query, key matrix respectively. Define the attention matrix as:
\begin{equation}\label{eq:attn_matrix_proof}
    Attn(Q,K):= \text{softmax}(QK^T/\sqrt{d}).
\end{equation}
\end{definition}

\begin{definition}\label{def:full_attn_proof}
    Let $Q, K, V\in\mathbb{R}^{L\times d}$ be the query, key matrix respectively. Define the full attention as:
\begin{equation}\label{eq:full_attn_proof}
    Attn_f(Q,K,V):= \text{softmax}(QK^T/\sqrt{d})V = Attn(Q,K)V.
\end{equation}
\end{definition}

\begin{definition}\label{def:win_attn_proof}
    Let $Q, K, V\in\mathbb{R}^{L\times d}$ be the query, key, value matrix respectively with $q_i, k_i, v_i$ the $i$-th row of the matrix $Q, K, V$. Let $w\in\mathbb{N}$ be the window size, such that $w$ divides $L$.  Define the window attention as:
    \begin{equation}\label{eq:win_attn_proof}
        Attn_w(Q, K, V, w):= 
        \begin{bmatrix}
             \sum_{j\in J(1)} \frac{\exp(q_1k_j^T/\sqrt{d})v_j}{\gamma_1}\\
            \vdots\\
            \sum_{j\in J(L)} \frac{\exp(q_Lk_j^T/\sqrt{d})v_j}{\gamma_L}
        \end{bmatrix}.
    \end{equation}
    Here $J(m) = \{Mw +1,\dots, (M+1)w\}$, where $M = \lfloor\frac{m-1}{w}\rfloor$. And \begin{equation}
    \gamma_m = \sum_{j\in J(m)} \exp(q_mk_j^T/\sqrt{d}).
\end{equation}

\end{definition}

\begin{definition}\label{def:mix_win_attn_matrix_proof}
    Let $A\in\mathbb{R}^{L\times L}$ and $Q, K, V, q_i, k_i, v_i, w$ be the same as defined in \textbf{Definition \ref{def:win_attn_proof}}, define the mixed window attention as:
\begin{equation}\label{eq:mix_win_attn_matrix_proof}
    Attn_{mw}(Q,K,V,w, A):= A Attn_w(Q,K,V,w)
\end{equation}
\end{definition}

\begin{theorem}\label{thm:full_attn=mw_attn_proof}
    Let $Q, K, V \in\mathbb{R}^{L\times d}$. Let $w\in \mathbb{N}$ such that $w$ divides $L$. If Attn($Q,K$) is BDI, then there exists a matrix $A\in \mathbb{R}^{L\times L}$ such that 
    \begin{equation}
        Attn_f(Q,K,V) = Attn_{mw}(Q,K,V,w, A).
    \end{equation}
    In particular, we can construct the exact value of $A$.
\end{theorem}


\begin{proof}
    We have 
    the $i$-th row of full attention is
\begin{equation}
    Attn_f^i = \sum_{j=1}^L \dfrac{\exp(q_i^Tk_j/\sqrt{d})v_j}{\beta_i},
\end{equation}
where $\beta_i = \sum_{j=1}^L \exp(q_ik_j^T/\sqrt{d})$. Let $\alpha_{im}$ be the $i,m$ entry of $A$, the $i$-th row of mixed window attention is
\begin{equation}
    Attn_{mw}^i = \sum_{m=1}^L \alpha_{im}\sum_{j=Mw+1}^{(M+1)w} \dfrac{\exp(q_m^Tk_j/\sqrt{d})v_j}{\gamma_m},
\end{equation}
where $\gamma_m = \sum_{j\in J(m)} \exp(q_mk_j^T/\sqrt{d})$, with $J(m) = \{Mw +1,\dots, (M+1)w\}$, where $M = \lfloor\frac{m-1}{w}\rfloor$.

Consider the following cases:
\begin{itemize}
    \item If $i=m$ and $j\in \{Mw+1,\dots, (M+1)w\}$, set $\dfrac{1}{\beta_i} = \dfrac{\alpha_{ii}}{\gamma_m}$.
    \item If $i\ne m$ and $j\in \{Mw+1,\dots, (M+1)w\}$ and $j\in \{Iw+1,\dots, (I+1)w\}$, where $I = \lfloor\frac{i-1}{w}\rfloor$, set $\alpha_{im} = 0$.
    \item If $i\ne m$ and $j \in \{Mw+1,\dots, (M+1)w\}$ and $j\not\in \{Iw+1,\dots, (I+1)w\}$, where $I = \lfloor\frac{i-1}{w}\rfloor$. For each $j$ we set
    \begin{equation}\label{eq:system_eqns_proof}
        \dfrac{\exp(q_ik_j^T/\sqrt{d})}{\beta_i} = \sum_{m\in J(j)} \dfrac{\alpha_{im} \exp(q_mk_j^T/\sqrt{d})}{\gamma_m}.
    \end{equation}
    For each $i$ and set of $\{m,j\}$ pairs, we have to solve a system of $w$ equations and unknowns. We now invoke the BDI assumption of $Attn(Q,K)$ to show that this system of equations has unique solution. 
    Let $C$ be the coefficient matrix of the right hand side of the system of equations in (\ref{eq:system_eqns_proof}). We observe that $C^T$ is invertible, because each row of $C^T$ is a scaled version of a square sub matrix along the diagonal of $Attn(Q,K)$, and each square sub-matrix along the diagonal of $Attn(Q,K)$ is invertible. The invertibility of $C$  then follows from that of $C^T$.
\end{itemize}
We completed the construction of $A$ as claimed in the theorem. 

\end{proof}

\textbf{Remark} The BDI assumption in \textbf{Theorem} \ref{thm:full_attn=mw_attn_proof} is a sufficient condition and not a necessary condition. We only need BDI to solve the system of equations in Eq. \ref{eq:system_eqns_proof}. Equation \ref{eq:system_eqns_proof} may be solvable even if BDI is not met. In this case, the solution will not be unique. In section \ref{sec:cond_num}, we will show in practice that BDI is satisfied by most of the datasets presented in this paper.

\begin{definition}\label{def:fwin_attn_proof}
    Let $A\in\mathbb{R}^{L\times L}$ and $Q, K, V, w$ be the same as defined in \textbf{Definition \ref{def:win_attn_proof}}, define the Fourier-mixed window attention as:
\begin{equation}\label{eq:fwin_attn_matrix_proof}
    Attn_{Fwin}(Q,K,V,w, A):= A \mathcal{F}(Attn_w(Q,K,V,w)),
\end{equation}
where $\mathcal{F}$ is the discrete Fourier transform.
\end{definition}

\begin{corollary}\label{cor:fwin_attn=full_attn_proof}
    Let $Q, K, V$ and $w$ be the same as defined in \textbf{Theorem \ref{thm:full_attn=mw_attn_proof}}. If Attn($Q,K$) is BDI, then there exists a matrix $A\in \mathbb{C}^{L\times L}$ such that 
    \begin{equation}
        \text{Attn}_f(Q,K,V) = \text{Attn}_{Fwin}(Q,K,V,w, A).
    \end{equation}
\end{corollary}
\begin{proof}
    From \textbf{Theorem \ref{thm:full_attn=mw_attn_proof}}, there exists $B\in\mathbb{R}^{L\times L}$ such that 
    \begin{equation}
    Attn_f(Q,K,V) = Attn_{mw}(Q,K,V,w, B).
    \end{equation}
    Thus if we let $A = B\mathcal{F}^{-1}$, then we are done.
\end{proof}

\begin{definition}\label{def:hwin_attn_proof}
    Let $A\in\mathbb{R}^{L\times L}$ and $Q, K, V, w$ be the same as defined in \textbf{Definition \ref{def:win_attn_proof}}, define the Hartley-mixed window attention as:
\begin{equation}\label{eq:hwin_attn_matrix_proof}
    \text{Attn}_{Hwin}(Q,K,V,w, A):= A \mathcal{H}(Attn_w(Q,K,V,w)),
\end{equation}
where $\mathcal{H}$ is the Hartley transform \cite{bracewell1986hartley}.
\end{definition}

\begin{corollary}\label{cor:hwin_attn=full_attn_proof}
    Let $Q, K, V$ and $w$ be the same as defined in \textbf{Theorem \ref{thm:full_attn=mw_attn_proof}}. If Attn($Q,K$) is BDI, then there exists a matrix $A\in \mathbb{R}^{L\times L}$ such that 
    \begin{equation}
        \text{Attn}_f(Q,K,V) = \text{Attn}_{Hwin}(Q,K,V,w, A).
    \end{equation}
\end{corollary}
\begin{proof}
    From \textbf{Theorem \ref{thm:full_attn=mw_attn_proof}}, there exists $B\in\mathbb{R}^{L\times L}$ such that 
    \begin{equation}
    Attn_f(Q,K,V) = Attn_{mw}(Q,K,V,w, B).
    \end{equation}
    Thus if we let $A = B\mathcal{H}^{-1}$, then we are done.
\end{proof}

\begin{table*}[ht!]
\caption{Standard deviation comparison on LSTF benchmarks (S/M for uni/multivariate), best results highlighted in bold.}
    \vskip 0.1in
    \centering
    \begin{tabular}{|c|c|c c|c c|||c c|c c|}
    \hline
         \multicolumn{2}{|c|}{Methods}&\multicolumn{2}{c|}{FWin (S)}&\multicolumn{2}{c|||}{Informer (S)}&\multicolumn{2}{c|}{FWin (M)}&\multicolumn{2}{c|}{Informer (M)} \\   
         \hline
         \multicolumn{2}{|c|}{Metric} & MSE & MAE & MSE & MAE & MSE & MAE & MSE & MAE \\
         \hline
         \multirow{5}{1em}{\rotatebox{90}{ETTh$_1$}} 
         & 24 & \textbf{0.012}& \textbf{0.019}& 0.022& 0.029& \textbf{0.011}& \textbf{0.008}& 0.058& 0.039\\
         & 48 & \textbf{0.019}& \textbf{0.027}& 0.036& 0.043& \textbf{0.043}& \textbf{0.028}& 0.115& 0.069\\
         & 168& \textbf{0.030}& \textbf{0.035}& 0.062& 0.063& 0.061& 0.030& \textbf{0.047}& \textbf{0.029}\\
         & 336& \textbf{0.012}& \textbf{0.016}& 0.057& 0.057& \textbf{0.033}& \textbf{0.013}& 0.142& 0.063\\
         & 720& \textbf{0.014}& \textbf{0.015}& 0.051& 0.055& 0.055& 0.019& \textbf{0.048}& \textbf{0.017}\\
         \hline
         \hline
         \multirow{5}{1em}{\rotatebox{90}{ETTh$_2$}} 
         & 24 & 0.005& 0.008& \textbf{0.004}& \textbf{0.006}& 0.043& 0.026& \textbf{0.028}& \textbf{0.021}\\
         & 48 & \textbf{0.010}& \textbf{0.011}& 0.015& 0.016& 0.140& 0.068& 0.239& 0.059\\
         & 168& 0.028& 0.024& \textbf{0.018}& \textbf{0.015}& \textbf{0.108}& \textbf{0.023}& 0.555& 0.108\\
         & 336& \textbf{0.022}& \textbf{0.019}& 0.036& 0.025& \textbf{0.169}& \textbf{0.044}& 0.407& 0.111\\
         & 720& \textbf{0.008}& \textbf{0.007}& 0.019& 0.013& \textbf{0.162}& \textbf{0.033}& 0.292& 0.050\\
         \hline
         \hline
         \multirow{5}{1em}{\rotatebox{90}{ETTm$_1$}} 
         & 24 & \textbf{0.002}& \textbf{0.009}& 0.009& 0.025& \textbf{0.012}& \textbf{0.007}& 0.045& 0.032\\
         & 48 & \textbf{0.006}& \textbf{0.013}& 0.020& 0.042& \textbf{0.021}& \textbf{0.010}& 0.037& 0.025\\
         & 96 & \textbf{0.016}& \textbf{0.023}& 0.040& 0.051& 0.060& \textbf{0.031}& \textbf{0.054}& 0.040\\
         & 288& \textbf{0.019}& \textbf{0.018}& 0.066& 0.054& \textbf{0.043}& \textbf{0.017}& 0.136& 0.070\\
         & 672& \textbf{0.033}& \textbf{0.039}& 0.083& 0.069& 0.102& 0.043& \textbf{0.058}& \textbf{0.028}\\
         \hline
         \hline
         \multirow{5}{1em}{\rotatebox{90}{Weather}} 
         & 24 & 0.006& \textbf{0.008}& \textbf{0.004}& 0.012& 0.010& 0.011& \textbf{0.003}& \textbf{0.001}\\
         & 48 & 0.010& 0.014& \textbf{0.008}& \textbf{0.009}& \textbf{0.003}& \textbf{0.001}& 0.005& 0.003\\
         & 168& 0.010& 0.013& \textbf{0.005}& \textbf{0.005}& 0.011& \textbf{0.006}& \textbf{0.006}& 0.014\\
         & 336& 0.035& 0.029& \textbf{0.011}& \textbf{0.011}& \textbf{0.012}& \textbf{0.007}& 0.015& \textbf{0.007}\\
         & 720& \textbf{0.011}& \textbf{0.009}& 0.025& 0.010& 0.015& 0.012& \textbf{0.011}& \textbf{0.007}\\
         \hline
         \hline
         \multirow{5}{1em}{\rotatebox{90}{ECL}} 
         & 48 & \textbf{0.011}& \textbf{0.010}& 0.015& 0.011& \textbf{0.011}& 0.010& \textbf{0.011}& \textbf{0.007}\\
         & 168& \textbf{0.006}& \textbf{0.005}& 0.017& 0.009& 0.006& \textbf{0.004}& \textbf{0.003}& \textbf{0.004}\\
         & 336& 0.030& 0.018& \textbf{0.007}& \textbf{0.009}& \textbf{0.004}& \textbf{0.003}& 0.046& 0.026\\
         & 720& 0.047& \textbf{0.021}& \textbf{0.032}& \textbf{0.021}& \textbf{0.007}& \textbf{0.007}& 0.090& 0.043\\
         & 960& \textbf{0.034}& \textbf{0.018}& 0.100& 0.054& \textbf{0.008}& \textbf{0.006}& 0.088& 0.044\\
         \hline
         \hline
         \multirow{4}{1em}{\rotatebox{90}{Exchange}} 
         & 96 & 0.071& \textbf{0.035}& \textbf{0.068}& 0.039& \textbf{0.115}& \textbf{0.045}& 0.130& 0.046\\
         & 192& \textbf{0.188}& \textbf{0.060}& 0.278& 0.083& 0.155& 0.052& \textbf{0.024}& \textbf{0.009}\\
         & 336& 0.162& 0.063& \textbf{0.138}& \textbf{0.032}& 0.065& 0.029& \textbf{0.053}& \textbf{0.016}\\
         & 720& 0.227& 0.078& \textbf{0.175}& \textbf{0.048}& \textbf{0.119}& \textbf{0.041}& 0.199& 0.051\\
         \hline
         \hline
         \multirow{4}{1em}{\rotatebox{90}{ILI}} 
         & 24& \textbf{0.126}& \textbf{0.035}& 0.211& 0.046& \textbf{0.137}& \textbf{0.031}& 0.142& 0.032\\
         & 36& 0.124& 0.037& \textbf{0.077}& \textbf{0.015}& \textbf{0.022}& \textbf{0.014}& 0.180& 0.039\\
         & 48& 0.148& 0.037& \textbf{0.057}& \textbf{0.017}& \textbf{0.044}& \textbf{0.011}& 0.168& 0.031\\
         & 60& \textbf{0.082}& \textbf{0.019}& 0.147& 0.027& \textbf{0.036}& \textbf{0.014}& 0.192& 0.036\\
         \hline
         \hline
         \multicolumn{2}{|c|}{Count}& \multicolumn{2}{c|}{42}& \multicolumn{2}{c|||}{23}& \multicolumn{2}{c|}{43} &  \multicolumn{2}{c|}{26}  \\
         \hline
    \end{tabular}

\label{tab:FWin_accuracy_std}
\vskip -0.1in
\end{table*}

\begin{table*}[ht!]
    \caption{Inference/training times (seconds) vs. prior-knowledge dependent (noted by *) transformers on univariate datasets.}
    \vskip 0.1in
    \centering
    \begin{tabular}{|c|c|c c|c c|c c|c c|}

    \hline
         \multicolumn{2}{|c|}{Methods}& \multicolumn{2}{c|}{Autoformer*}& \multicolumn{2}{c|}{FEDformer*} &\multicolumn{2}{c|}{ETSformer*} & \multicolumn{2}{c|}{FWin} \\
         \hline
         \multicolumn{2}{|c|}{Metric}& Inference & Train & Inference & Train & Inference & Train & Inference & Train\\
         \hline
         \multirow{5}{1em}{\rotatebox{90}{ETTh$_1$}} 
         & 24 & 0.7056 & 242.70 & 0.7039 & 379.09 & 0.1766 & 113.76 & 0.1123 & 267.57\\
         & 48 & 0.6717 & 276.59 &  0.7530 & 431.62& 0.1987 & 191.99 & 0.1720 & 528.90  \\
         & 168 & 0.8726 & 365.88 & 0.7715 & 548.20& 0.1972 & 351.06 & 0.1804 & 548.34 \\
         & 336 & 0.9025 & 526.16 & 0.7582 & 695.77& 0.3041 & 736.35 & 0.1794 & 545.03\\
         & 720 & 1.0161 & 860.87 & 0.7857 & 971.53& 0.2889 & 677.74 & 0.1760 & 516.01\\
         \hline
         \hline
         \multirow{5}{1em}{\rotatebox{90}{ETTm$_1$}} 
         & 24 & 0.6795 & 924.45 & 0.6956 & 1421.72& 0.1782 & 561.69 & 0.1015 & 281.33   \\
         & 48 & 0.6926 & 1024.84 & 0.6699 & 1632.67& 0.1770 & 580.23 & 0.0991 & 310.58   \\
         & 168 & 0.7326 & 1119.16 & 0.8113 & 2133.04& 0.1786 & 593.16 & 0.1076 & 1065.35\\
         & 288 & 0.9187 & 1981.56 & 0.8088 & 2546.94& 0.1832 & 803.27 & 0.1714 & 2203.90\\
         & 672 & 1.0137 & 3659.89 & 0.8631 & 3963.63& 0.1870 & 1163.27 & 0.1776 & 2284.08\\
         \hline
         \hline

         \multirow{5}{1em}{\rotatebox{90}{ECL}} 
         & 48 & 0.6846 & 477.14 & 0.7202 & 890.71 & 0.2191 & 824.47 & 0.1091 & 558.92\\
         & 168 & 0.8439 & 641.90 & 0.7641 & 1163.12 & 0.2142 & 908.73 & 0.1171 & 626.27  \\
         & 336 & 0.8577 & 956.78 & 0.7767 & 1404.01 & 0.2180 & 941.34 & 0.1143 & 587.11  \\
         & 720 & 1.0372 & 1654.55 & 0.7695 & 2067.92 & 0.2732 & 1048.18 & 0.1804 & 1128.26\\
         & 960 & 1.2184 & 2548.90 & 0.7963 & 2419.27 & 0.2949 & 1058.18 & 0.2054 & 1171.78 \\
         \hline
    \end{tabular}

    \label{tab:Univariate_Train_Inference_Time_Appendix}
    \vskip -0.1in
    \end{table*}

\begin{table*}[ht!]
    \caption{Inference/training times (seconds) vs. prior-knowledge dependent (noted by *) transformers on multivariate datasets.}
    \vskip 0.1in
    \centering
    \begin{tabular}{|c|c|c c|c c|c c|c c|}
    \hline
         \multicolumn{2}{|c|}{Methods}& \multicolumn{2}{c|}{Autoformer*} & \multicolumn{2}{c|}{FEDformer*}& \multicolumn{2}{c|}{ETSformer*} & \multicolumn{2}{c|}{FWin}  \\
         \hline
         \multicolumn{2}{|c|}{Metric} &Inference & Train & Inference & Train & Inference & Train & Inference & Train \\
         \hline
         \multirow{5}{1em}{\rotatebox{90}{ETTh$_1$}} 
         & 24 & 0.6884 & 220.29 & 0.6706 & 359.67 & 0.1758 & 122.99 & 0.0996 & 50.83 \\
         & 48 & 0.7778 & 248.73 & 0.7083 & 413.29 & 0.1749 & 190.34 & 0.1018 & 73.34 \\
         & 168 & 0.9023 & 369.76 & 0.8082 & 565.03 & 0.2009 & 329.34 & 0.1059 & 156.38  \\
         & 336 & 0.8885 & 530.38 & 0.7562 & 711.03 & 0.2752 & 702.21 & 0.1098 & 195.80  \\
         & 720 & 1.0892 & 866.92 & 0.7403 & 997.00 & 0.2854 & 694.10 & 0.1791 & 557.36 \\
         \hline
         \hline
         \multirow{5}{1em}{\rotatebox{90}{ETTm$_1$}} 
         & 24 & 0.6878 & 842.76 & 0.7036 & 1335.71 & 0.1777 & 542.67 & 0.1075 & 1029.65 \\
         & 48 & 0.6994 & 899.90 & 0.7148 & 1546.38 & 0.1831 & 606.23 & 0.1162 & 317.65  \\
         & 168 & 0.7765 & 1135.78 & 0.7582 & 2192.51 & 0.1853 & 695.64 & 0.1120 & 1102.62 \\
         & 288 & 0.9372 & 2014.46 & 0.8598 & 2591.67 & 0.1805 & 954.43 & 0.1835 & 2690.85 \\
         & 672 & 1.0269 & 3654.97 & 0.7848 & 3989.08 & 0.1799 & 1341.47 & 0.1960 & 2761.32\\
         \hline
         \hline

         \multirow{5}{1em}{\rotatebox{90}{ECL}} 
         & 48 & 0.6518 & 555.99 & 0.7089 & 1010.28 & 0.2566 & 1459.48 & 0.1245 & 290.75  \\
         & 168 & 0.8908 & 938.61 & 0.9080 & 1667.01 & 0.2482 & 1664.97 & 0.1205 & 696.57 \\
         & 336 & 0.9236 & 1466.83 & 0.9468 & 2297.83 & 0.2696 & 1927.64 & 0.1377 & 1297.49\\
         & 720 & 1.1052 & 2608.97 & 0.8800 & 3592.86 & 0.3195 & 2566.90 & 0.2273 & 3352.05 \\
         & 960 & 1.2642 & 4298.18 & 0.9097 & 4473.98 & 0.3396 & 2858.18 & 0.2463 & 3445.76 \\
         \hline
    \end{tabular}

    \label{tab:Multivariate_Train_Inference_Time_Appendix}
    \vskip -0.1in
\end{table*}

\section{Additional Experimental Data and Details}\label{sec:exp_details}
\textbf{Exchange} \footnote{
https://github.com/laiguokun/multivariate-time-series-data} \cite{lai2018modeling}: The dataset contains daily exchange rates of eight different countries from 1990 to 2016. The train/val/test split ratio is 7:1:2.

\textbf{ILI} \footnote{
https://gis.cdc.gov/grasp/fluview/fluportaldashboard.html}: The dataset contains weekly recorded influenza-like illness (ILI) patients data from Centers for Disease Control and Prevention of the United States from 2002 to 2021. This describes the ratio of patients seen with ILI and the total number of the patients. The train/val/test split ratio is 7:1:2.

\textbf{Traffic} \footnote{
https://pems.dot.ca.gov}: The dataset contains hourly data from California Department of Transportation, describing the road occupancy rates measured by different sensors on San Francisco Bay area freeways. The train/val/test split ratio is 7:1:2.
Results on the traffic data are in Tab. \ref{tab:traffic_accuracy} where FWin are comparable to (much better than) Informer on univariate (multivariate) data. 
The relative MSE difference between Informer and FWin on univariate (multivariate) data is 5\% (12.23\%). 
The relative MAE difference between Informer and FWin on univariate (multivariate) data is 
5.89\% (11.68\%).


\subsection{Setup of experiments}
For all the experiments to compute the errors, the encoder's input sequence and the decoder's start token are chosen from $\{24, 48, 96, 168, 336, 720\}$ for the ETTh$_1$, ETTh$_2$, Weather and ECL dataset, and from $\{24, 48, 96, 192, 288, 672\}$ for the ETTm dataset. The default window size is 24. We use a window size of 12 when the encoder's input sequence is 24. For the Exchange, ILI, and Traffic datasets, we use the same hyper-parameters as those provided in Autoformer. The encoder's input sequence is 96 and decoder's input sequence is 48. The window size is 24 for Exchange and Traffic. For ILI, we use an input length of 36 for the encoder and 18 for decoder, with window size of 6. The number of windows on the cross attention is set to 3. The models are trained for 6 epochs with learning rate adjusted by a factor of 0.5 every epoch. 

For the power grid dataset \cite{powergrid_toolbox,glassoformer_22}, the encoder and decoder inputs are set to 200. The prediction length is 700, and the window size is 25. The models are trained for 80 epochs with an early stopping counter of 30. The learning rate is adjusted every 10 epochs by a factor of 0.85.

\subsection{Setup of train/inference time}\label{subsec:fwin_vs_informer_time}
The results in Tab. \ref{tab:Train_Inference_Time} are obtained by conducting experiments using the same set of hyper-parameters for each model and dataset. The inference time represents the total duration taken by a model to generate 30 predictions. The training time is a run of simulation each consisting of 6 epochs. Run time is very sensitive to machine conditions, thus we ran all of the simulations using the same machine, under the same traffic conditions. 

\section{FWin Accuracy with Standard Deviation}
In Tab. \ref{tab:FWin_accuracy_std}, we include the standard deviations of the FWin and Informer accuracies shown in the paper. The standard deviations are computed using five experimental runs.  

\section{Inference and Training Time Comparison with Auto/Fed/ETSformers}
We compare inference and training times of 
FWin transformer with those of Autoformer \cite{autoformer_21}, FEDformer \cite{fedformer} and ETSformer \cite{woo2022etsformer} in Tab. \ref{tab:Univariate_Train_Inference_Time_Appendix} and Tab. \ref{tab:Multivariate_Train_Inference_Time_Appendix}.
These three transformers are designed to improve accuracies of Informer with prior-knowledge of datasets. e.g. using trend/seasonality decomposition. 
In contrast, FWin has no prior-knowledge based operation, exactly the same as Informer. 
 We use the default set of hyper-parameters provided in their respective papers.
 The inference time represents the total duration taken by a model to generate 30 predictions. The training time is a run of simulation each consisting of 6 epochs.




\section{Condition Number of Attention Matrix}
\label{sec:cond_num}
Theorem \ref{thm:full_attn=mw_attn} requires the attention matrix to be BDI. In this section, we will verify that in practice BDI is satisfied by many of the datasets here. The experimental set-up is:
\begin{itemize}
    \item Run simulations on the Informer model using full attention instead of probsparse.
    \item Collect the full attention matrix of the first encoder block of the Informer.
    \item Given a window size $w$, compute the condition numbers of  $w\times w$ sub-matrices along the diagonal of the attention matrix. 
    \item If the condition numbers are finite, then BDI is true and this instance is collected for a histogram plot, otherwise it is counted as a failure. 
\end{itemize}
Using the procedure above, we plot all condition numbers for all simulations of the datasets in the paper. For all the simulations, we use the same hyper-parameters as in the experiment sections. Due to memory space constraint, we report the first 11 batches of the test datasets. 

\begin{figure*}[ht!]
    \centering
    \includegraphics[width=0.8\textwidth]{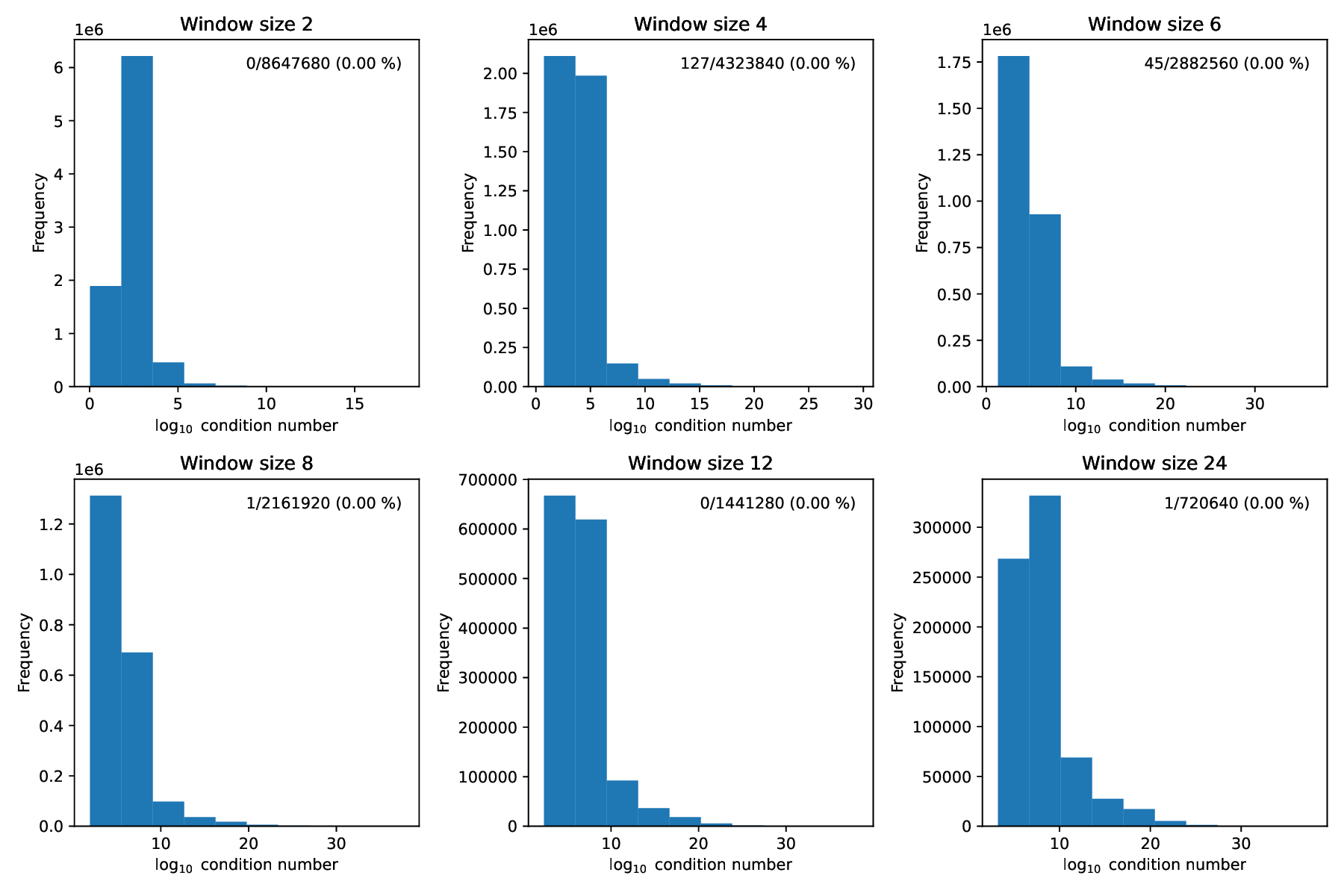}
    \caption{Condition number of ETTh1 (M) dataset under various window sizes. On the top right corner of each subplot there is a label ``n/m (k \%)", here m denotes the total number of condition numbers, n denotes the number of condition numbers that are infinite, and $k$ denote the percentage of condition numbers that is infinite.}
    \label{fig:all_cond_num_ETTh1}
\end{figure*}

\begin{figure*}[ht!]
    \centering
    \includegraphics[width=0.8\textwidth]{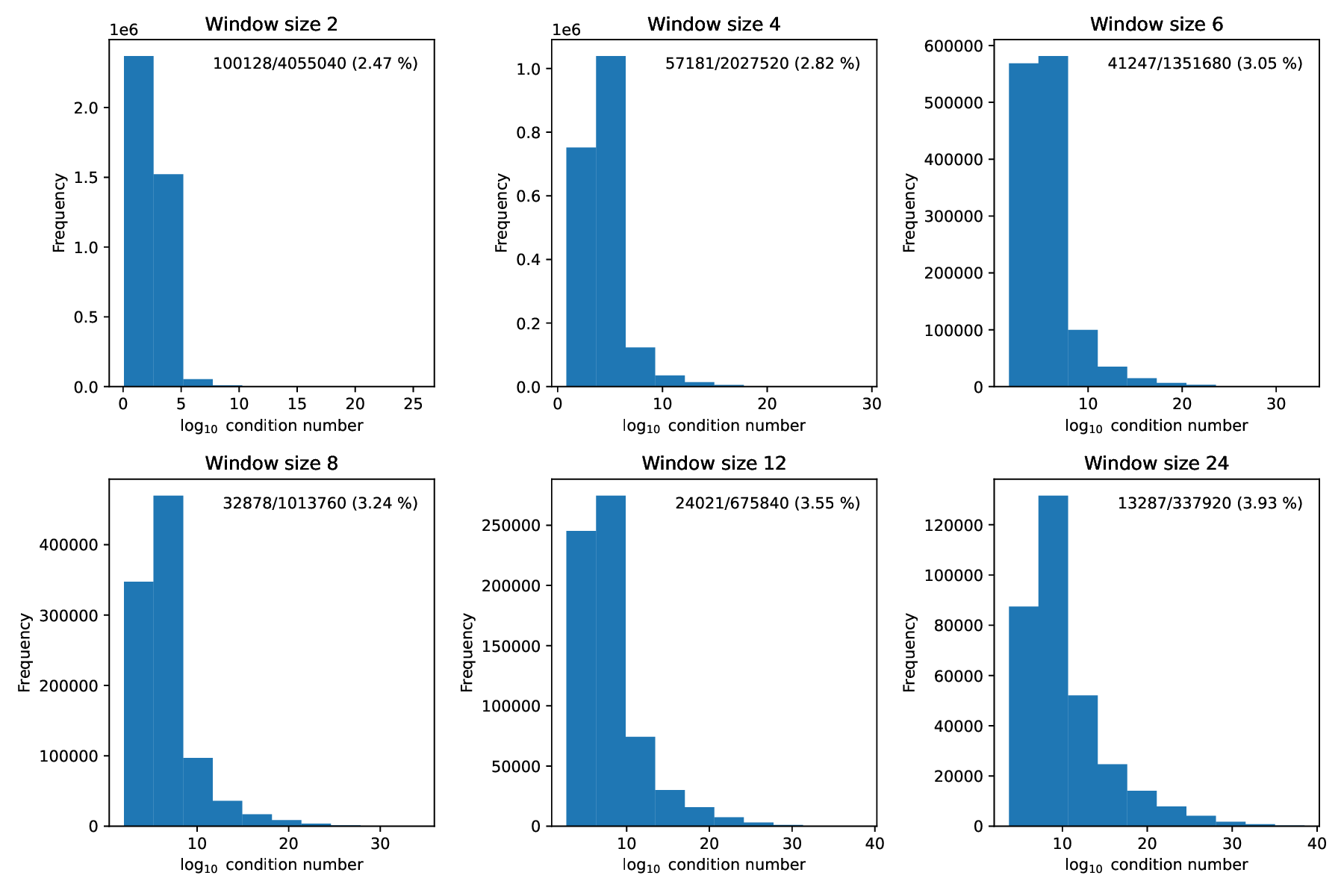}
    \caption{Condition number of Weather (M) dataset under various window sizes. On the top right corner of each subplot there is a label ``n/m (k \%)", here m denotes the total number of condition numbers, n denotes the number of condition numbers that are infinite, and $k$ denote the percentage of condition numbers that is infinite.}
    \label{fig:all_cond_num_WTH}
\end{figure*}

\begin{figure*}[ht!]
    \centering
    \includegraphics[width=0.8\textwidth]{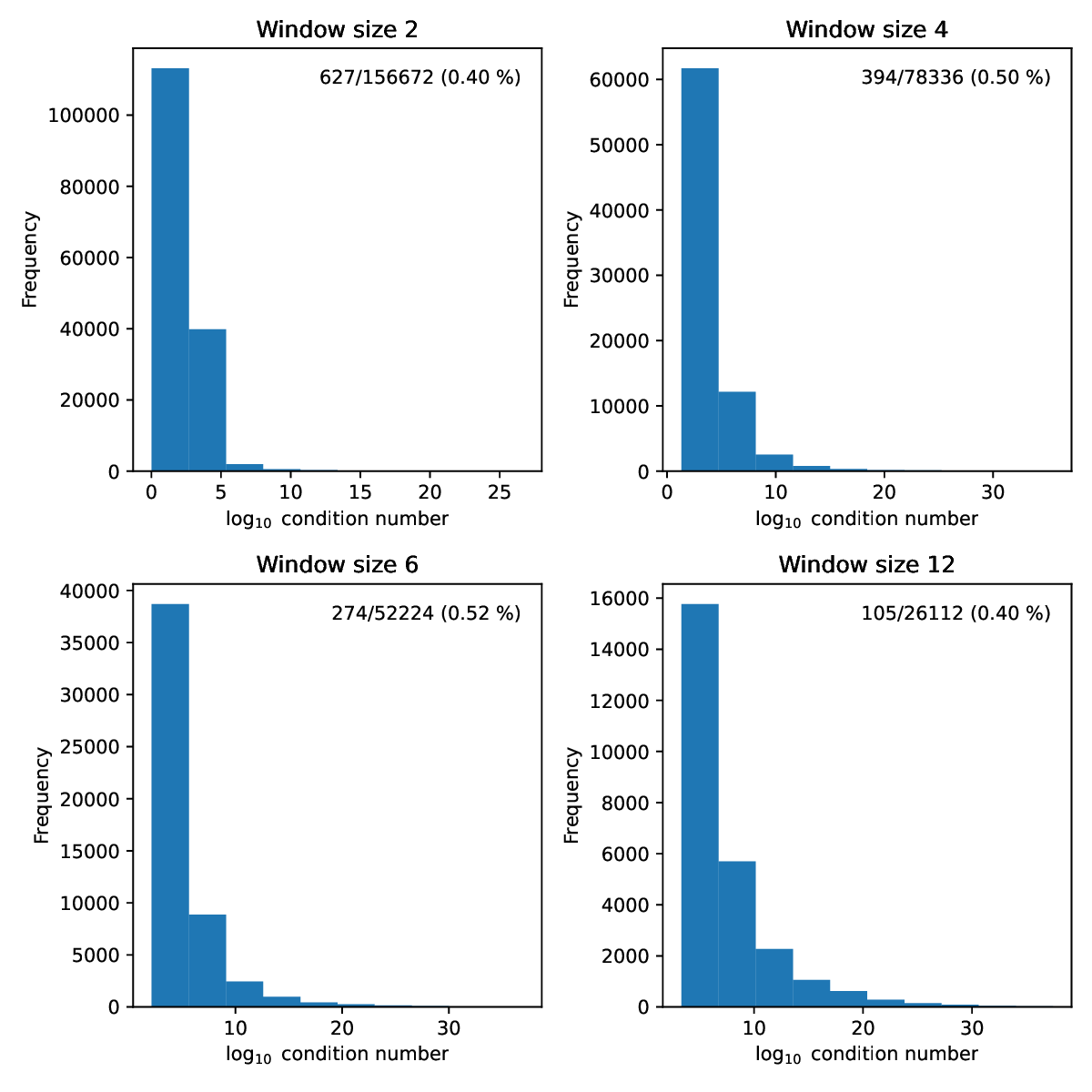}
    \caption{Condition number of ILI (M) dataset under various window sizes. On the top right corner of each subplot there is a label ``n/m (k\%)", here m denotes the total number of condition numbers, n denotes the number of condition numbers that are infinite, and $k$ denote the percentage of condition numbers that is infinite.}
    \label{fig:all_cond_num_ILI}
\end{figure*}

From Figs. \ref{fig:all_cond_num_ETTh1}, \ref{fig:all_cond_num_WTH}, \ref{fig:all_cond_num_ILI}, we observe that ETTh1, Weather, ILI datasets satisfy the assumption of the theorem relatively well. In particular, ETTh1 and ILI do not have many infinite condition numbers. The Weather dataset has about 3\% infinite condition numbers. We also noted that the condition numbers increase with the window size. 

\section{FWin and Nonparametric Regression }\label{sec:nonparam}

The full self-attention function  \cite{vaswani2017attention} can be conceptualized as an estimator in a non-parametric kernel regression problem in statistics \cite{nguyen2022fourierformer}. 
Let the key vectors serve as the training inputs and the value vectors as the training targets. 
The key-value pairs $\{k_j,v_j\}$ for $j=1,\dots,N$, come from the model
\begin{equation}\label{regmod}
    v_j = f(k_j) + \epsilon_j, 
\end{equation}
where $f$ is an unknown function to be estimated and $\epsilon_j$ are zero mean independent noisy perturbations. Let the key vectors $k_1, k_2, \dots, k_N$ be i.i.d. samples from a distribution function $p(k)$, and 
the key-value pairs $(v_1,k_1),\dots, (v_N, k_N)$ be i.i.d. samples from the joint density $p(k,v)$. Since $\mathbb{E}[v_j |k_j] = f(k_j)$, 
the classical Nadaraya-Watson method \cite{Nad_64,Paz_62,Ros_56} approximates $p$ by a sum of Gaussian kernels and gives the estimate of $f$ below:
\begin{align}
   \hat{f}_\sigma(k) := \sum_{j=1}^N \dfrac{v_j\phi_\sigma(k-k_j)}{\sum_{j=1}^N \phi_\sigma(k-k_j)},
\end{align}
where $\phi_\sigma(\cdot)$ is the isotropic multivariate Gaussian density function with diagonal covariance matrix $\sigma^2\mathbf{I}_D$. In particular if $k=q_i$, and the $k_j$'s are normalized, one obtains 
$ \hat{f}_\sigma(q_i)=$
\begin{equation}\label{eqn:attn_approx}
  \dfrac{\sum_{j=1}^N v_j \exp(q_ik_j^T/\sigma^2)}{\sum_{j=1}^N\exp(q_ik_j^T/\sigma^2)}
  =\sum_{j=1}^N \text{softmax}(q_ik_j^T/\sigma^2)v_j.
\end{equation}
Letting $\sigma^2 = \sqrt{d_\text{model}}$, where $d_\text{model}$ is the dimension of $q_i$ ($k_j$), turns estimator  (\ref{eqn:attn_approx}) into the softmax self-attention (\ref{eqn:attn_eqn}). 

To build a window attention, we allow the query vector $q_i$ to interact only with nearby key and value vectors. Thus the window version of the softmax estimator is
 $\bar{f}_\sigma(q_i):=$
\[ \dfrac{\sum_{j\in J(i)} v_j \exp(q_ik_j^T/\sigma^2)}{\sum_{j\in J(i)}\exp(q_ik_j^T/\sigma^2)} 
 =\sum_{j\in J(i)} \text{softmax}(q_ik_j^T/\sigma^2)v_j  
 \]
where $J(i)$ is the index set that correspond to the set of keys the query $q_i$ interact with. 
In view of the fully connected (MLP) layer after the Fourier mixing layer and before the output in Fig. \ref{fig:model_overview},  we define the analogous FWin estimator for the regression model as
\begin{equation}
    \tilde{f}_\sigma(q_i) := A \cdot \mathcal{R}(\mathcal{F}(\bar{f}_\sigma(q_i))),
\end{equation}
where $\mathcal{R}$ takes the real part, $\mathcal{F}$ is the discrete Fourier transform, $\cdot$ represents matrix multiplication, and $A$ is a real 
matrix to be learned from the training data 
by minimizing the sum of squares error (MSE) 
of the regression model (\ref{regmod}).

\begin{figure}[ht!]
    \centering
    \includegraphics[width=\columnwidth]{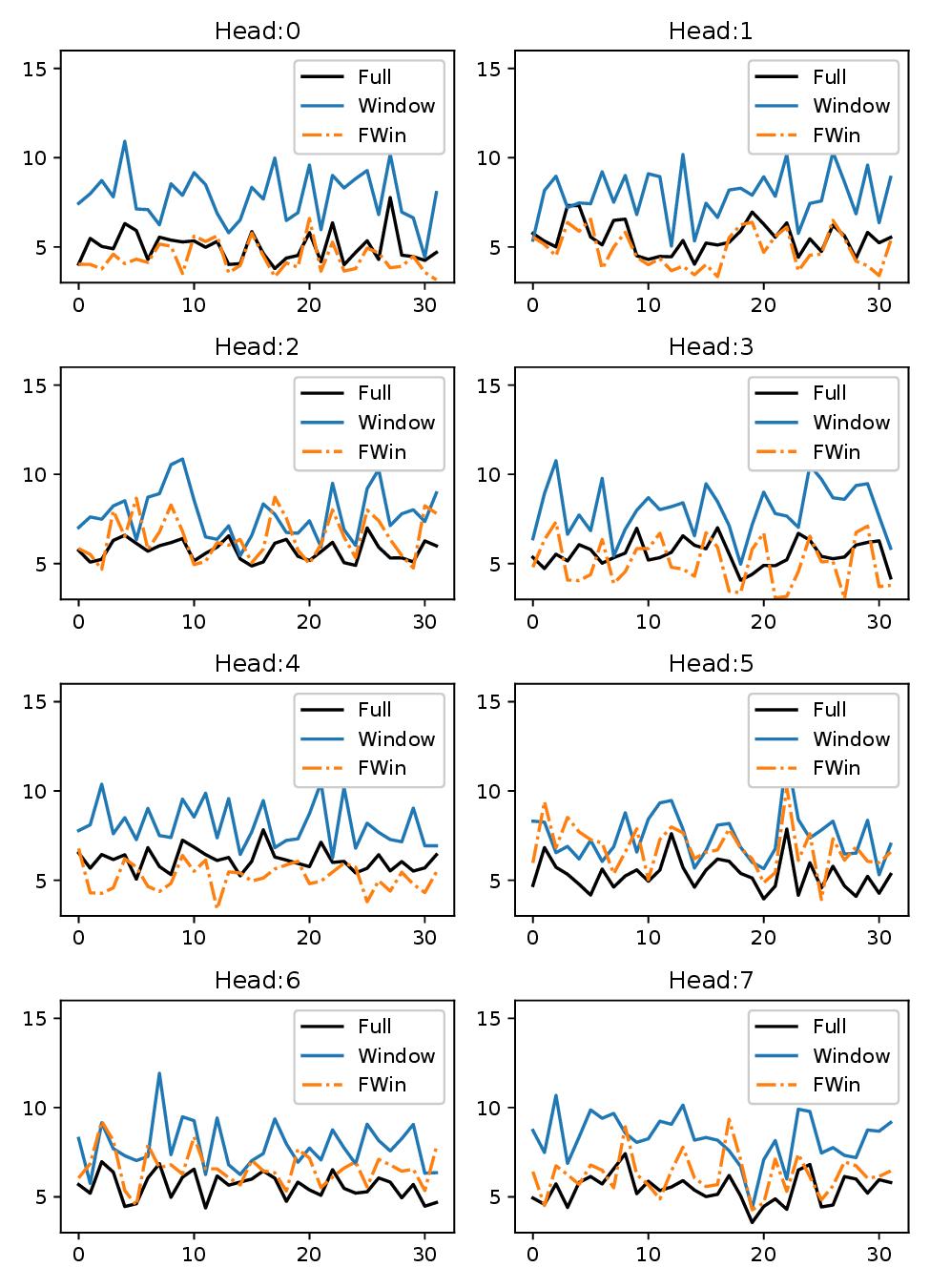}
    \caption{MSE error vs. query number comparison of full softmax attention (black), window attention (blue) and FWin attention (orange) in the non-parametric regression model (\ref{regmod}) based on key vectors of a full attention layer of Informer$^{\dagger}$ \cite{informer_21} trained from the ETTh$_1$ data set.}
    \label{fig:nonparametric_learn}
\end{figure}

{\bf Kernel Regression Experiment}
To examine the differences among the three  estimators $\hat{f}$, $\bar{f}$, and $\tilde{f}$, we opt for the Laplace distribution function $f = \exp(-\alpha|x|)$, for $\alpha=0.01$, as the ground truth nonlinear function acting componentwise on the input to the regression model (\ref{regmod}). We use a set of query, key, value vectors from 
Informer$^\dagger$ \cite{informer_21}
on ETTh$_1$ multivariate data set with prediction length (metric) of 720. We choose this particular data set because Informer$^\dagger$ \cite{informer_21} with full softmax self-attention 
has the best performance there. 
The key vectors may not satisfy the theoretical i.i.d. assumption \cite{nguyen2022fourierformer}. 
In this experiment, we have a set of $168$ query and key vectors from each of the 8 heads. Denote query $q_i$, and key $k_j$ vectors for $i, j = 1, 2, \dots, 168$. The value vectors are $v_j = f(k_j)$. We divide the data into 136 vectors for training and the remaining 32 for testing. The mean square error (MSE) in testing of the estimator $\hat{f}(q_{i_n})$ for $n=1,2,\dots,32$, are labeled full estimator in Fig. \ref{fig:nonparametric_learn}. Similarly, we compute MSEs for  $\bar{f}_\sigma(q_{i_n})$, and $\tilde{f}(q_{i_n})$, using window size of 4, where $A$ is  learned by solving a least squares problem. Fig. \ref{fig:nonparametric_learn} compares  the MSE of the three estimators over data from 8 heads. We observe that the FWin estimator consistently outperforms the window attention estimator, and approaches the full softmax attention. In heads 0/1/4, FWin outperforms the full softmax attention estimator which is not theoretically optimal for the regression task \cite{nguyen2022fourierformer}. 
In conclusion, the 
regression experiment on the three estimators indicates that 
FWin is a simple and reliable local-global attention structure with competitive capability, lending added support to 
its robust performance.

\newpage
\bibliography{bib}
\bibliographystyle{icml2024}
\end{document}